\pdfoutput=1

\documentclass[10pt,journal,compsoc]{IEEEtran}
\ifCLASSOPTIONcompsoc
  \usepackage[nocompress]{cite}
\else
  \usepackage{cite}
\fi
\ifCLASSINFOpdf
  \usepackage[pdftex]{graphicx}
\else
\fi
\usepackage{amsmath}
\usepackage{algorithmic}

\usepackage{array}
\graphicspath{{figures/}{pictures/}{images/}{./}} 

\usepackage{hyperref}
\usepackage{cite}                      
\usepackage{tabu}                      
\usepackage{booktabs}                  
\usepackage{paralist}
\usepackage{enumitem}
\usepackage[most]{tcolorbox}
\usepackage{xcolor}
\usepackage{wrapfig}

\usepackage{tikz}
\usepackage{amsmath}
\usepackage{amssymb}
\usepackage[linesnumbered,ruled,vlined]{algorithm2e}
\usepackage{graphicx,calc}
\newlength\myheight
\newlength\mydepth
\settototalheight\myheight{Xygp}
\newcommand*\inlinegraphics[1]{%
  \settototalheight\myheight{Xygp}%
  \settodepth\mydepth{Xygp}%
  \raisebox{-\mydepth}{\includegraphics[height=\myheight]{#1}}%
}

\definecolor{maroonred}{rgb}{0.69, 0.137, 0.094}
\definecolor{royalblue}{rgb}{0.259, 0.455, 0.694}

\definecolor{darkgreen}{rgb}{0.0, 0.7, 0.0}

\definecolor{colorfour}{HTML}{EC008C} 
\definecolor{colorfive}{HTML}{00B0F0}
\definecolor{colornine}{HTML}{009B55}
\newcommand{\clrfour}{\textcolor{colorfour}}
\newcommand{\clrnine}{\textcolor{colornine}}
\newcommand{\clrfive}{\textcolor{colorfive}}

\newcommand{\headfour}{\clrfour{\textit{layer 0 head 4}}}
\newcommand{\headnine}{\clrnine{\textit{layer 0 head 9}}}
\newcommand{\headfive}{\clrfive{\textit{layer 11 head 5}}}

\newcommand*\circled[1]{\tikz[baseline=(char.base)]{
            \node[shape=circle,draw,inner sep=1pt] (char) {#1};}}

\newcommand{\imageoverview}{\textit{Image Overview}}

\newcommand{\headimportance}{\textit{Head Importance View}}

\newcommand{\attnstrength}{\textit{Attention Strength View}}

\newcommand{\attnpattern}{\textit{Attention Pattern View}}

\definecolor{ronecolor}{rgb}{0.937, 0.745, 0.173}

\newtcbox{\ronebox}{on line,
  colframe=white,        
  colback=ronecolor,
  coltext=white,        
  boxrule=0.1pt,        
  arc=2pt,              
  boxsep=0.1pt,
  left=2pt,right=2pt,top=1pt,bottom=1pt,
}

\definecolor{rtwocolor}{rgb}{0.518, 0.671, 0.314}

\newtcbox{\rtwobox}{on line,
  colframe=white,        
  colback=rtwocolor,
  coltext=white,        
  boxrule=0.1pt,        
  arc=2pt,              
  boxsep=0.1pt,
  left=2pt,right=2pt,top=1pt,bottom=1pt,
}

\definecolor{rthreecolor}{rgb}{0.447, 0.6, 0.824}

\newtcbox{\rthreebox}{on line,
  colframe=white,        
  colback=rthreecolor,
  coltext=white,        
  boxrule=0.1pt,        
  arc=2pt,              
  boxsep=0.1pt,
  left=2pt,right=2pt,top=1pt,bottom=1pt,
}

\SetCommentSty{mycommfont}
\SetKw{Continue}{continue}
\SetKwProg{Fn}{def}{\string:}{}
\SetFuncArgSty{textnormal}
\SetArgSty{textnormal}

\begin{document}
%
\title{How Does Attention Work in Vision Transformers? A Visual Analytics Attempt}

%
%
%
%

\author{Yiran~Li,
        Junpeng~Wang,
        Xin~Dai,
        Liang~Wang,
        Chin-Chia~Michael~Yeh,
        Yan~Zheng, \\
        Wei~Zhang,
        and~Kwan-Liu~Ma,~\IEEEmembership{Fellow,~IEEE}
\IEEEcompsocitemizethanks{\IEEEcompsocthanksitem Y. Li and K.-L. Ma are with University of California, Davis.\protect\\
E-mail: \{ranli, klma\}@ucdavis.edu
\IEEEcompsocthanksitem J. Wang, X. Dai, L. Wang, C.-C. M. Yeh, Y. Zheng, and W. Zhang are with Visa Research, Palo Alto, CA, 94301.\protect\\
E-mail: \{junpenwa, xidai, liawang, miyeh, yazheng, wzhan\}@visa.com
}

\thanks{Manuscript received October 28, 2022; revised February 24, 2023.}}

%
%

\markboth{Journal of \LaTeX\ Class Files,~Vol.~14, No.~8, August~2015}%
{Shell \MakeLowercase{\textit{et al.}}: Bare Demo of IEEEtran.cls for Computer Society Journals}
%



\IEEEtitleabstractindextext{%
\begin{abstract}
Vision transformer (ViT) expands the success of transformer models from sequential data to images. The model decomposes an image into many smaller patches and arranges them into a sequence. Multi-head self-attentions are then applied to the sequence to learn the attention between patches. 
Despite many successful interpretations of transformers on sequential data, little effort has been devoted to the interpretation of ViTs, and many questions remain unanswered. For example, among the numerous attention heads, which one is more important? 
How strong are individual patches attending to their spatial neighbors in different heads? What attention patterns have individual heads learned? 
In this work, we answer these questions through a visual analytics approach. Specifically, we first identify \textbf{\textit{what}} heads are more important in ViTs by introducing multiple pruning-based metrics. 
Then, we profile the spatial distribution of attention strengths between patches inside individual heads, as well as the trend of attention strengths across attention layers.
Third, using an autoencoder-based learning solution, we summarize all possible attention patterns that individual heads could learn. Examining the attention strengths and patterns of the important heads, we answer \textbf{\textit{why}} they are important. 
Through concrete case studies with experienced deep learning experts on multiple ViTs, we validate the effectiveness of our solution that deepens the understanding of ViTs from \textit{head importance}, \textit{head attention strength}, and \textit{head attention pattern}.
\end{abstract}

}

\maketitle

\IEEEdisplaynontitleabstractindextext

%
\IEEEpeerreviewmaketitle


%
%
%
%

\IEEEraisesectionheading{\section{Introduction}\label{sec:introduction}}
\IEEEPARstart{T}{ransformer} models have demonstrated outstanding performance on tasks in natural language processings (NLP)~\cite{vaswani2017attention,devlin2019bert,liu2019roberta} and time-series forecastings~\cite{lim2021temporal}. Recently, their success has also been extended to the vision domain, and the resulting vision transformer (ViT) has achieved on-par and even better performance than the state-of-the-art CNNs~\cite{dosovitskiy2020image}.
ViT converts a 2D image into a 1D sequence by decomposing it into many patches and arranging the patches sequentially. Each patch is analogous to a token of sequential data, and the multi-head self-attentions are then applied to the sequence to learn the relation among tokens.

Despite the superb performance, it remains unclear how ViT works internally, especially how the multi-head self-attention works on image patches. To be specific, we found ViT designers are often puzzled by the following questions: 

\begin{itemize}[leftmargin=12pt, topsep=0pt,itemsep=0pt,parsep=0pt,partopsep=0pt]
\item \textit{First}, how important are individual heads, and is their importance consistent across images? 
As different heads emphasize different pair-wise attentions, their contributions to the prediction are also different. Identifying the important ones would limit the scope of model analysis. 

\item \textit{Second}, how strong is the attention between two patches that are nearby or far away from each other, and does the attention strength show any trend across layers?
It is widely known that CNNs extract basic shapes/colors in early layers but complex objects/concepts in later layers. Since ViTs demonstrate on-par performance with CNNs, it becomes a natural question to ask if the models have any learning heuristic from early to later layers.

\item \textit{Third}, what attention patterns have individual heads learned, and are those patterns related to image contents? We have observed heads with interesting patterns, e.g., always attending to the patch itself regardless of the image content (content-agnostic) or only attending to patches with target objects (content-relevant). But, there lacks an exhaustive summary of all possible patterns.
\end{itemize}

\noindent Answering these questions will provide a fundamental understanding of ViTs and assist their further development.

\begin{figure*}
\centering
 \includegraphics[width=\textwidth]{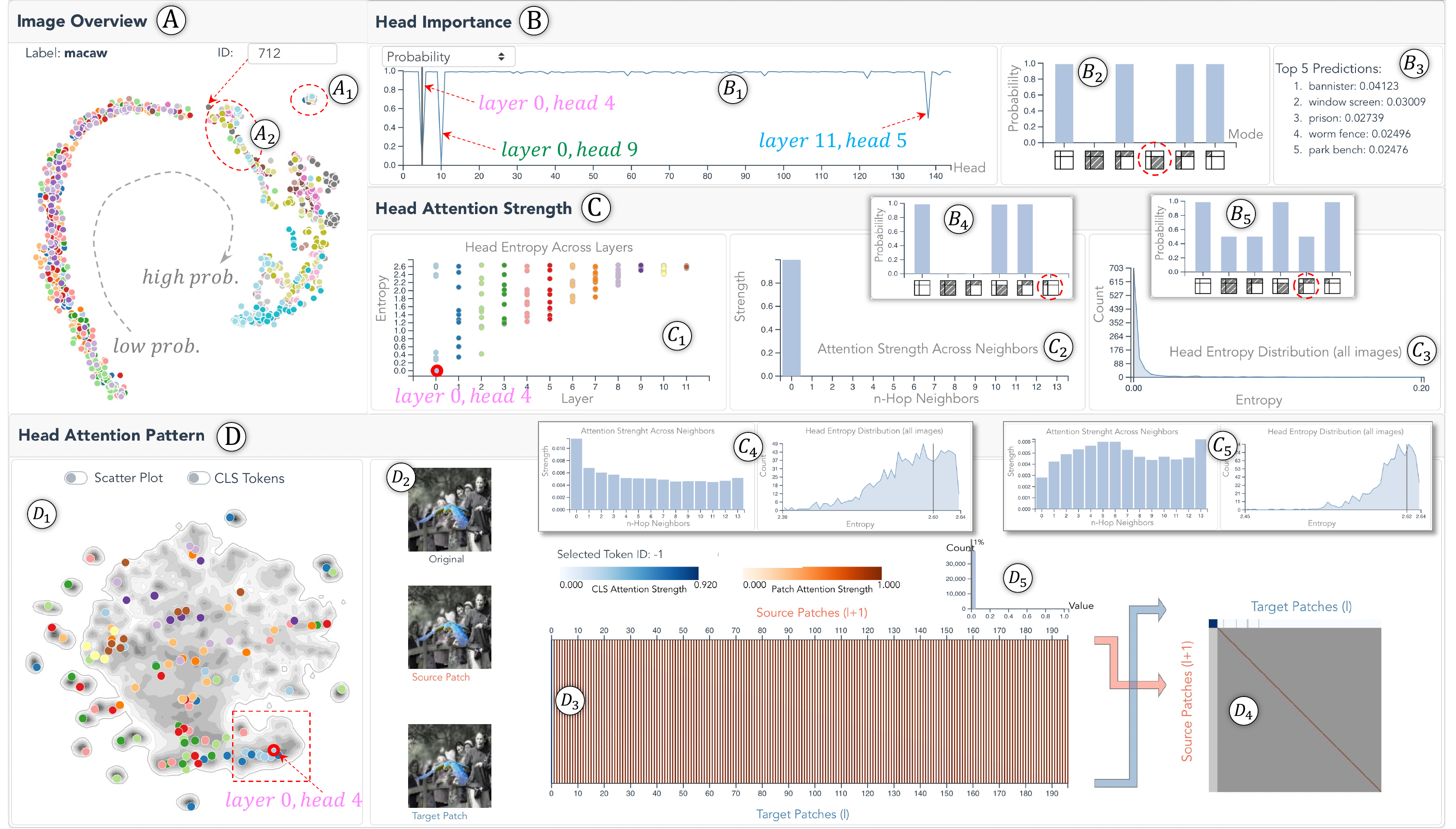}
 \vspace{-0.3in}
 \caption{Our system contains four components. The \imageoverview{} (A) lays out all images for selection. The \headimportance{} (B) shows all heads in different importance metrics (B1) and dissects a head's importance through partial pruning (B2, B3). The attention strengths between patches in a head are shown in the \attnstrength{} (C), where users first obtain an overview of all heads (C1), and then focus on one head (C2, C3). 
 The \attnpattern{} (D) clusters all heads by their attention pattern (D1), and presents the pattern details in one head (D2${\sim}$D5).}
 \label{fig:teaser}
\end{figure*}

However, there are multiple challenges in answering them.
\textit{First}, Michel and Levy~\cite{michel2019aresixteen} proved that heads are not equally important in transformers through intensive ablation studies. Hao et al.~\cite{hao2021self} proposed a self-attention attribution score for each head to quantify its importance through token interactions. These works focus on NLP tasks and use a head's impact on the prediction to verify its importance. In ViTs, however, we find that heads with little impact on the ultimate predictions may considerably influence the intermediate representations, indicating the need to profile head importance from multiple perspectives. 
\textit{Second}, for attention strengths and attention patterns, multiple works have proposed to visualize them with heatmap, flow map, or matrix visualizations~\cite{kovaleva2019revealing, park2019sanvis, derose2020attention,li2021t3}. However, all these works focus on language transformers with 1D attentions (forward/backward). ViTs, though rearrange image patches into 1D, still present 2D attention behaviors in the 2D spatial context (e.g., upward/downward attention). 
The 2D attention behavior results in much richer attention patterns, and the patches' attention strength to their neighbors needs to be redefined considering the spatial distance in 2D.
\textit{Lastly}, based on our collaborations with domain experts, existing ViT analyses are often piece-by-piece and lack a fluent analytical workflow. For example, to understand the attention between image patches, the patches and attentions between them need to be presented intuitively and explored coordinately. However, most ViT researchers still connect the two parts manually by eyeballing them back and forth.

Our work interprets ViTs from three aspects: \textit{head importance}, \textit{head attention strengths}, and \textit{head attention patterns}.
For head importance, we introduce multiple pruning-based head importance metrics, which are computed offline and can be easily plugged into our system to support head exploration and importance analysis.
For attention strengths, we profile a patch's attention strength to its $k$-hop neighbors as a $k$-dimensional vector. Aggregating the vectors from all patches of a head reflects the head's attention strength distribution. For attention patterns, we train an autoencoder for the heads' attention matrices, and summarize all possible attention patterns by clustering the latent representations of all heads. The three parts are integrated into a coordinated visual analytics (VA) system. We validate the system's efficacy by studying different ViTs with experienced deep learning experts. In short, our contributions include:
\begin{enumerate}[leftmargin=15pt, topsep=0pt,itemsep=0pt,parsep=0pt,partopsep=0pt]
    \item Multiple pruning-based metrics describing ViT heads' importance from different perspectives.
    \item A characterization of heads' attention strength across image patches' $k$-hop neighbors.
    \item A comprehensive summary of the possible attention patterns in ViTs using an autoencoder-based solution.
    \item An interactive visual analytics system integrating the above three parts for coordinated interpretations of ViTs.
\end{enumerate}

\section{Related Work}
Our work belongs to the visual analytics attempts towards more interpretable deep learning (DL), with a special focus on interpreting multi-head self-attention from transformers. We thus review earlier works from these two aspects.

\textit{\textbf{Visualizations for DL.}} 
A plethora of visualization works have been introduced for the interpretation of deep neural networks recently~\cite{liu2016towards,strobelt2017lstmvis,wang2018ganviz,wang2018dqnviz,jin2022gnnlens}.
We refer readers to recent surveys~\cite{hohman2018visual, choo2018visual} for a thorough review of these works. Lately, deep transformers demonstrate superior performance than other DL models on 1D sequential data, and multiple visualization works have been introduced for their interpretations~\cite{kovaleva2019revealing,park2019sanvis,vig2019multiscale,jaunet2021visqa, derose2020attention,li2021t3}. The success of transformers has also been extended to 2D images with the seminal work of vision transformers (ViTs)\cite{dosovitskiy2020image}. However, to the best of our knowledge, no comprehensive visual analyses have been conducted to demystify this type of powerful yet complex models, especially how attention works in the 2D image context. Our work tries to fill this gap.

\textbf{\textit{Attention Visualization.}} 
The attention mechanism~\cite{bahdanau2014neural} has been used extensively in DL, especially NLP-related tasks, to learn what target tokens the source tokens should ``look at''. Essentially, attention is a matrix where each cell denotes the attention magnitude that the source token (row) pays to the target (column). Popular attention visualization techniques include flow maps~\cite{dong2020interactive, strobelt2018s}, parallel coordinates plots (PCPs)~\cite{vig2019multiscale}, and heatmaps~\cite{jaunet2021visqa, park2019sanvis,aflalo2022vl}. For example, the flow maps used by Dong et al.~\cite{dong2020interactive} connect the source and target tokens with curves, the widths of which denote the attention strengths. 
Vig~\cite{vig2019multiscale} arranges the source and target tokens along two parallel axes (i.e., a simplified PCP) and connects them with line segments in between to show the attention patterns.  
Heatmaps are used extensively in NLP~\cite{abnar2020quantifying,kovaleva2019revealing,li2021t3}, where the attention strengths are directly encoded into the color of each heatmap cell. 
There are also customized visual designs for attention visualizations~\cite{derose2020attention, jaunet2021visqa}. For instance, DeRose et al.~\cite{derose2020attention} extract an ``attention graph'' from the attentions across layers of a BERT model and arrange the graph into a radial layout to present the propagation of attentions layer-by-layer. The resulting visualization, named Attention Flows, helps to easily analyze and compare attentions from two transformer models. 

For \textit{attention patterns} (in individual transformer heads), researchers have discovered some typical ones~\cite{park2019sanvis}, analyzed their occurrence in different tasks~\cite{kovaleva2019revealing}, 
compared the patterns between low and high-performing models~\cite{jin2022visual}, and related them with the corresponding heads' importance~\cite{li2021t3}. 
However, these works all focus on 1D sequential data. Attentions learned from images with a 2D spatial context have much richer patterns that are difficult to be identified and summarized manually. Our work intends to efficiently discover them.
For \textit{attention strengths} between patches within a ViT head, \textit{mean attention distance} has been introduced in previous works~\cite{cao2020behind, dosovitskiy2020image, raghu2021vision}, which is a sum of the attentions between patches weighted by their spatial distance. This single aggregated value holistically reflects each head's attention strength, but also averages out many spatial details. Here, we introduce the \textit{attention strength vector} to comprehensively profile the spatial distribution of attention strengths.

\section{Background}
\label{sec:background}

\begin{figure}[tbh]
 \centering 
 \includegraphics[width=0.9\columnwidth]{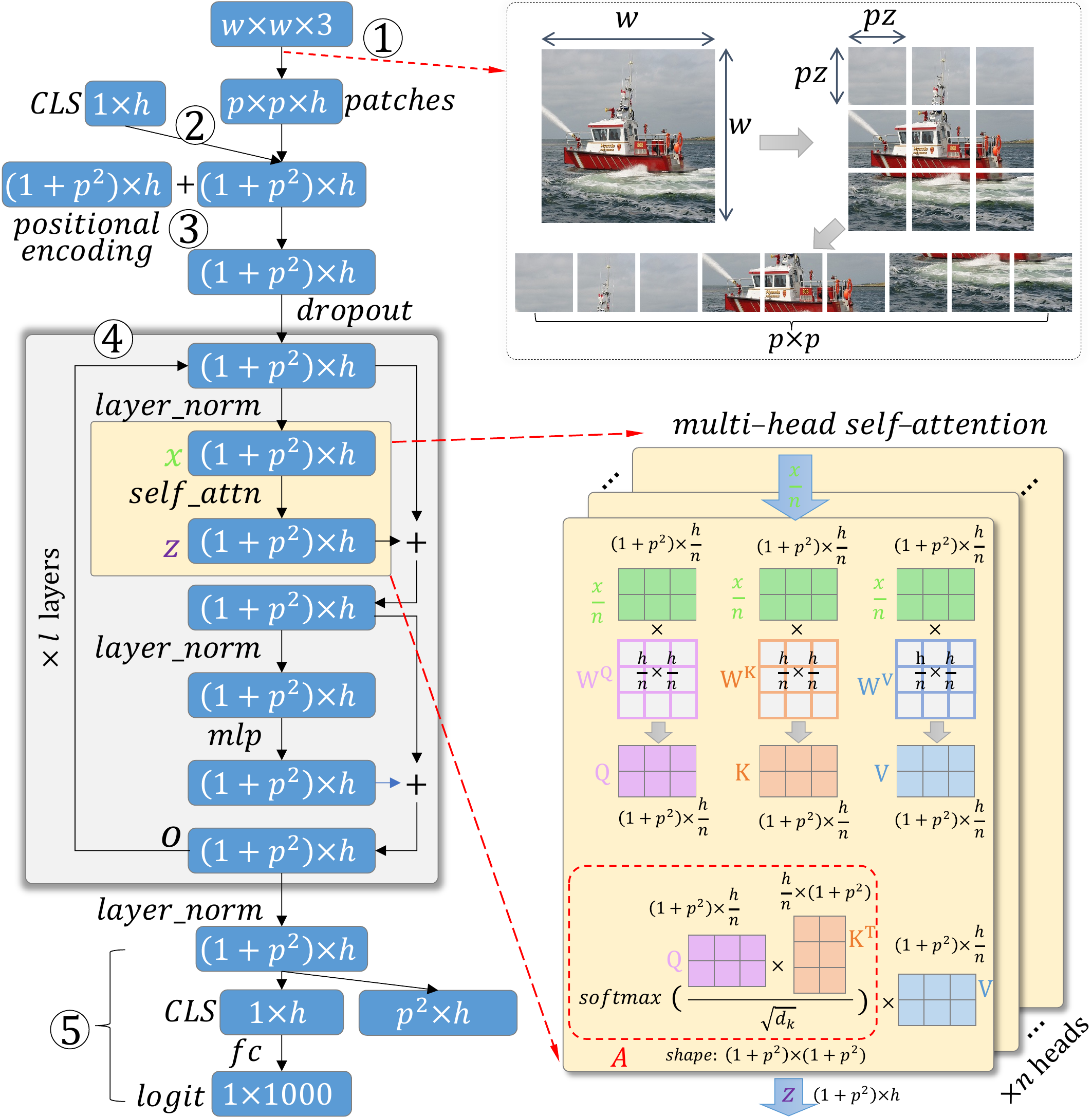}
  \vspace{-0.12in}
 \caption{ViT executes in five steps: (1) decompose the input image into patch tokens; (2) concatenate the \texttt{CLS} token; (3) add the positional encoding; (4) multi-layer multi-head self-attention; (5) use \texttt{CLS} for prediction. Step (4) includes $l$ attention layers, each has $n$ heads. The attention weight in each head, i.e., $A$, is our interpretation focus. }
 \label{fig:architecture}
\end{figure}

\textbf{ViT Model.} The most popular ViT application is image classification, which is also the focus of this paper. As shown in Fig.~\ref{fig:architecture}\circled{1}-\circled{5}, a ViT classifier runs in five key steps:

\begin{enumerate}[leftmargin=15pt, topsep=0pt,itemsep=0pt,parsep=0pt,partopsep=0pt]
\item \textit{Decompose the input image into a sequence of patch tokens}. Without loss of generality, we assume the same width and height for each input RGB image, denoted as $w$. If the patch size is $pz{\times}pz$, the number of patches will be $p^2{=}\frac{w}{pz}{\times}\frac{w}{pz}$. The patches are then arranged into a sequence of tokens; each is encoded as an $h$-dimensional ($h$D) vector. Each patch token learns a concise representation for the corresponding image patch.

\item \textit{Concatenate \texttt{CLS}.} A zero-initialized $h$D class token (\texttt{CLS}) is concatenated with the $p^2$ patch tokens, resulting in a $(1{+}p^2){\times}h$ matrix. \texttt{CLS} learns to accumulate class-related features used to generate the final class probability.

\item \textit{Add positional encodings.} The zero-initialized positional encodings are added to the $(1{+}p^2){\times}h$ matrix. They are trained to learn each patch's positional information. We skip their details as they are not our interpretation focus.

\item \textit{Multi-head self-attention.} This step contains $l$ stacked attention layers, each with $n$ attention heads. Each head learns a $(1{+}p^2){\times}(1{+}p^2)$ attention weight matrix $A$, 
reflecting the pair-wise attention between all $1{+}p^2$ tokens.

\item \textit{Use the \texttt{CLS} token for prediction.} This step decouples the \texttt{CLS} embedding from the patch tokens, and transforms it into class logits through fully-connected layers.  

\end{enumerate}

\textbf{Self-Attention.} Step 4 is the most important. The self-attention in each attention head (one yellow slice in Fig.~\ref{fig:architecture}, right) gathers information from all $1{+}p^2$ tokens to learn how much attention each token should pay to itself and others. The attentions are then used to update the tokens' representations. Specifically, the $(1{+}p^2){\times}h$ matrix at the end of Step 3, after some dropout and normalization layers, is evenly split over the $n$ heads, each with the shape of $(1{+}p^2){\times}\frac{h}{n}$. Inside each head, the matrix is further transformed into $Q$, $K$, and $V$ through three separate learnable weight matrices $W^Q$, $W^K$, and $W^V$. The self-attention is then computed as:
\vspace{-0.1in}
\begin{equation}\small
\label{eq:single_head}
Attention(Q, K, V) = A{\cdot}V = softmax(\frac{Q K^{T}}{\sqrt{d_K}}){\cdot}V.
\end{equation}

\noindent The attention weight $A$ of size $(1{+}p^2){\times}(1{+}p^2)$ encodes the pair-wise attention between all $1{+}p^2$ tokens. For clarity, we call the $1{+}p^2$ tokens \textit{\textbf{source}} tokens when they attend to others, but \textit{\textbf{target}} tokens when they are attended by others.

\textbf{Multi-Layer and Multi-Head.} The self-attention computation is conducted in all $n$ heads, and the resulting attentions are concatenated and linearly transformed to generate the final multi-head self-attention, denoted as $z$, i.e.,
\vspace{-0.05in}
\begin{equation}\small
\label{eq:multi_head}
z= Concat(head_1, head_2, \dots, head_n){\cdot}W^O + b,
\end{equation}
where $head_i{=}Attention(Q_i, K_i, V_i)$. As shown in 
Fig.~\ref{fig:architecture}\circled{4}, $z$ will go through more layers to generate the final attention layer output $o$ 
with shape $(1{+}p^2){\times}h$, which is the updated $h$D representation for the $1{+}p^2$ tokens. It will be fed to a new self-attention layer, and the process is repeated for $l$ times, resulting in $l$ stacked layers. In total there are
$l{\times}n$ attention heads, each with an attention weight matrix $A$ recording the learned attention between patches in the respective heads.

\section{Requirements and Solution Overview}
\label{sec:requirement}

We maintained weekly discussions with five domain experts (all are full-time researchers with 5+ years of deep learning experience) working on transformers in vision, NLP, and time-series domains. Over these discussions and our review of the tasks in related literature~\cite{derose2020attention, park2019sanvis, michel2019aresixteen,hao2021self}, we elicit the following design requirements for a visual analytics system.

\noindent
\ronebox{R1}: \textbf{\textit{Head Importance.}} To start the interpretation, we first need to quantify the importance of a large number of heads and dissect their importance. Specifically, this requires us to:

\begin{itemize}[leftmargin=15pt, topsep=0pt,itemsep=0pt,parsep=0pt,partopsep=0pt]
    
\item \textbf{R1.1}: 
Assess the importance of a ViT head. We want to reflect a head's impact on both its own attention layer and the ViT's final predictions. The impact should be assessed both on a single image and over all images.

\item \textbf{R1.2}: Dissect a head's importance. This is to disclose the contributions from two types of tokens to a head's importance: the \texttt{CLS} learns class-related features for prediction; the patch tokens learn important image contents. 

\item \textbf{R1.3}: 
Use head importance to guide image explorations. 
To analyze important heads, users need to select the right images for which the corresponding heads show importance. Therefore, we need to provide an informative overview of a large number of images based on their head importance to guide their exploration.

\end{itemize}

\noindent
\rtwobox{R2}: \textbf{\textit{Head Attention Strength.}} From the original ViT paper~\cite{dosovitskiy2020image} and the domain experts, we noticed that most ViT designers are wondering how the patches distribute their attention strengths spatially, e.g., whether they attend more to near/far patches and how the attention strength distribution is different across heads. Thus, we should answer:

\begin{itemize}[leftmargin=15pt, topsep=0pt,itemsep=0pt,parsep=0pt,partopsep=0pt]
    \item \textbf{R2.1}: For a single head of an image, how strong are patches attending to their spatially near/far neighbors?
    
    \item \textbf{R2.2}: For all heads of an image, does their attention show any patterns across layers? What are the patterns?
   
    \item \textbf{R2.3}: For a single head, does its attention strength show consistent spatial distributions across all images?
\end{itemize}

\noindent
\rthreebox{R3}: \textbf{\textit{Head Attention Pattern.}} As ViT shows more and much richer attention patterns in the 2D context, it is crucial to disclose them with the image semantics. Thus, we need to:
\begin{itemize}[leftmargin=15pt, topsep=0pt,itemsep=0pt,parsep=0pt,partopsep=0pt]
    \item \textbf{R3.1}: Exhaustively summarize all possible attention patterns from the $l{\times}n$ heads (for both \texttt{CLS} and patch tokens) and provide an effective overview of the patterns.
     
    \item \textbf{R3.2}: Drill down to individual heads of an image to effectively present its attention pattern and investigate if it is agnostic/relevant to the image contents.

\end{itemize}

\begin{figure}[tb]
    \centering
    \includegraphics[width=.9\columnwidth]{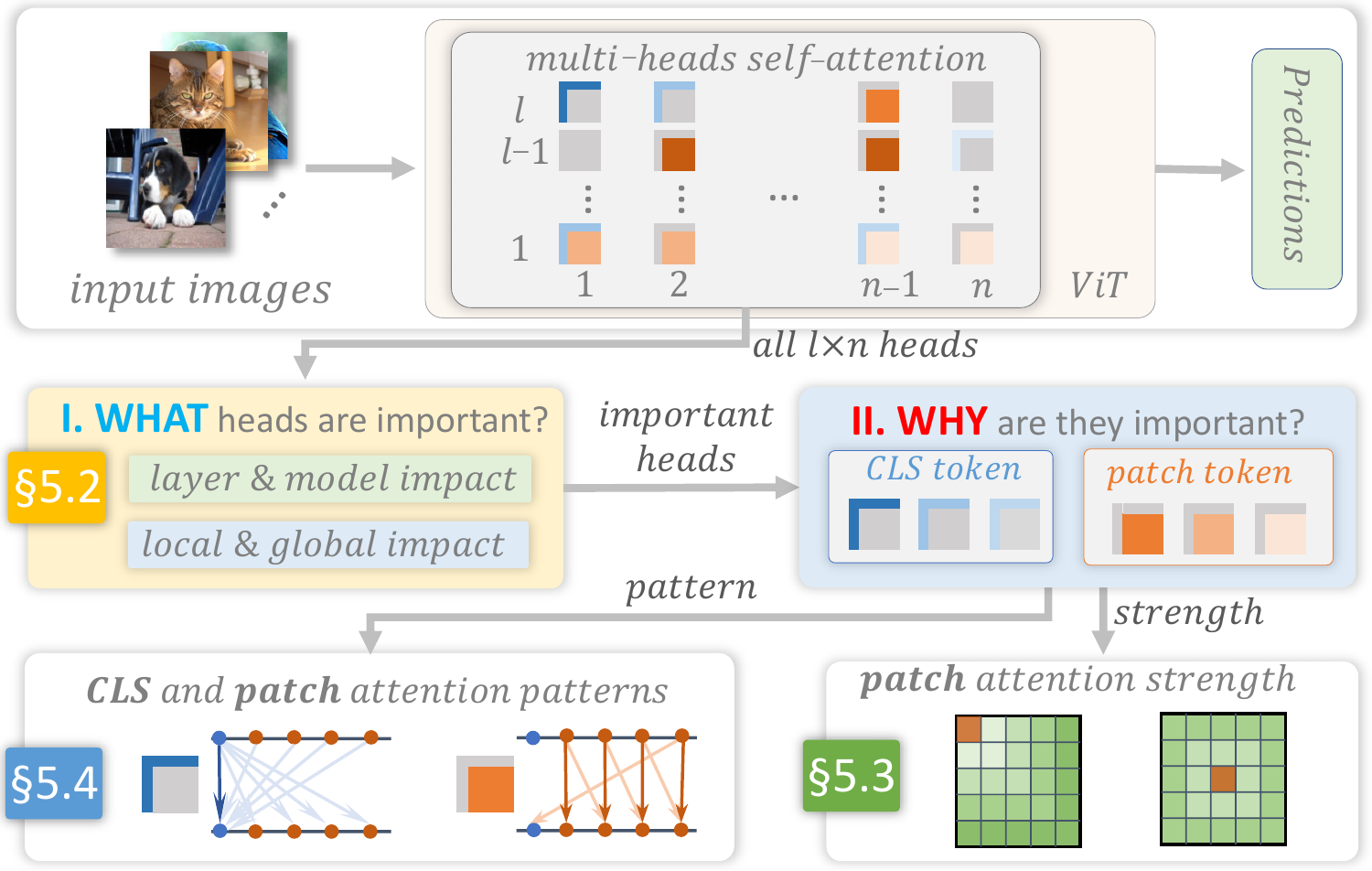}
    \vspace{-0.15in}
    \caption{Overview of our visual interpretation solution for ViT.}
    \label{fig:analysis-workflow}
\end{figure}

\textbf{Solution Overview.} We design a visual analytics system to meet the above requirements. Fig.~\ref{fig:analysis-workflow} illustrates the system's workflow.
To interpret a well-trained ViT, we first feed the image of interest into the model to get its $l{\times}n$ heads. Next, we answer \textit{what} heads are important (Fig.~\ref{fig:analysis-workflow} I) through four pruning-based metrics, meeting \ronebox{R1}. Focusing on these important heads, we explain \textit{why} they are important (Fig.~\ref{fig:analysis-workflow} II) from two perspectives. First, we disclose the attention strength distribution in individual heads by averaging attention strength across $k$-hop neighbors of individual patches (\rtwobox{R2}). Second, using an unsupervised clustering method, we summarize the attention patterns in both \texttt{CLS} and patch tokens and visualize the patterns in important heads (\rthreebox{R3}). An integrated visualization system (Fig.~\ref{fig:teaser}) has been developed following the workflow.
\setlength{\belowcaptionskip}{-10pt}
\setlength{\belowcaptionskip}{0pt}

\section{Methodology and Visualization System}
\label{sec:system}

Our visual analytics system (Fig.~\ref{fig:teaser}) contains four components: the \imageoverview{}, the \headimportance{}, the \attnstrength{}, and the \attnpattern{}. The \imageoverview{} (Fig.~\ref{fig:teaser}A) lays out image instances based on their heads' importance vector, providing an entry point for the exploration. The remaining three views (Fig.~\ref{fig:teaser}B-D) are designed to meet the three requirements.

\subsection{The Image Overview}
The \imageoverview{} (Fig.~\ref{fig:teaser}A) uses tSNE+scatterplot to provide an overview of the images. 
Each point represents an image, and its color denotes the class label. The coordinates of each point are the dimensionality reduction result of the corresponding image's \textit{head importance vector}, i.e., a $(l{\times}n)$-dimensional vector with the corresponding head's importance (Eq.~\ref{eq:prob}) at each dimension.
The tSNE layout based on this vector clusters images with similar head importance together, guiding users' exploration (\textbf{R1.3}). For example, there is a small cluster in the top-right corner, which immediately catches users' attention during exploration (see Sec.~\ref{sec:case_study}). 

Clicking on any point or providing an ID in the top-right input box will select the corresponding image into the other three views. Inside each view, the analysis can focus on the selected image or be extended to all other images.

\subsection{The Head Importance View}
\label{sec:head_importance}
We define several metrics to quantify the importance of a head. 
These metrics are generated by ``leave-one-out'' ablations, i.e., encoding a head's importance by the changes in the final output (\textit{model-level impact}) or next-layer activations (\textit{layer-level impact}) after pruning the head. Pruning a head is conducted by setting its attention matrix ($A$ in Eq.~\ref{eq:single_head}) to 0.
Similar head/neuron importance analysis through ablation studies has been widely adopted in NLP, e.g.,~\cite{bau2018identifying,michel2019aresixteen,hao2021self}.

\subsubsection{Importance to the Model's Output (\textbf{R1.1})}
\label{sec:global_importance}
We propose two \textit{model-level} importance metrics for each head.
One reflects the probability change of the true class (Eq.~\ref{eq:prob}); the other encodes the Jensen-Shannon Divergence (JSD) between the two probability distributions (Eq.~\ref{eq:jsd}) before and after a head is pruned.
Mathematically, $ViT()$ denotes the well-trained model, which takes an image as input and outputs a probability distribution, i.e., $\textbf{P} {=} ViT(img)$. $ViT_{i, j}()$ is the same model but the $j$th head from the $i$th layer has been pruned, and $\textbf{P}_{i,j} {=}ViT_{i, j}(img)$. $idx_{label}$ is the image's true class index. The importance of head $(i, j)$ is:
\begin{equation}\small
\label{eq:prob}
   I^{prob}_{i,j} = \textbf{P}{[idx_{label}]}-\textbf{P}_{i,j}[idx_{label}]
\end{equation}
\vspace{-0.15in}
\begin{equation}\small
\label{eq:jsd}
   I^{JSD}_{i,j} = JSD(\textbf{P}||\textbf{P}_{i,j}) 
\end{equation}

\subsubsection{Importance to the Attention Layer (\textbf{R1.1})}
\label{sec:local_importance}
Assessing only the changes in final outputs cannot reflect a head's importance in its attention layer. As our experts noticed, pruning an important head may significantly change the corresponding layer's output (i.e., $z$ in Eq.~\ref{eq:multi_head}), but show minor changes to the final probabilities. This is because heads from later layers may compensate for the contribution of the pruned head, concealing its importance.

To identify the important heads in each attention layer, we propose two \textit{layer-level} importance metrics, which are defined by the cosine distance ($D_{cos}$) between the immediate layer activations before and after a head is pruned. As shown in Fig.~\ref{fig:architecture}, the attention layer's output $z$ is a $(1{+}p^2){\times}h$ matrix, containing the activations of the \texttt{CLS} (the first $1{\times}h$) and patch tokens (the later $p^2{\times}h$). Our layer-level metrics measure the importance of the \texttt{CLS} and patch tokens separately.
For \texttt{CLS}, the metric ($I_{i,j}^\texttt{CLS}$) reflects the cosine distance between the two \texttt{CLS} activations. For patch tokens, the metric ($I_{i,j}^{patch}$) similarly computes the cosine distances and averages the distances over all patches.
Mathematically ($z$ and $z'$ are the layer's output before and after pruning),
\begin{equation}\small
    I_{i,j}^{{CLS}} = D_{cos}(z[0], z'[0])
    \label{eq:cls}
\end{equation}
\begin{equation}\small
    I_{i,j}^{patch} =\frac{1}{p^2}\Sigma_{i=1}^{p^2} D_{cos}(z[i], z'[i])
    \label{eq:patch}
\end{equation}

\subsubsection{Head Pruning Modes (\textbf{R1.2})}

Once the important heads are identified, we further dissect their importance by partially pruning them.
\begin{figure}[tbh]
    \centering
    \includegraphics[width=.88\columnwidth]{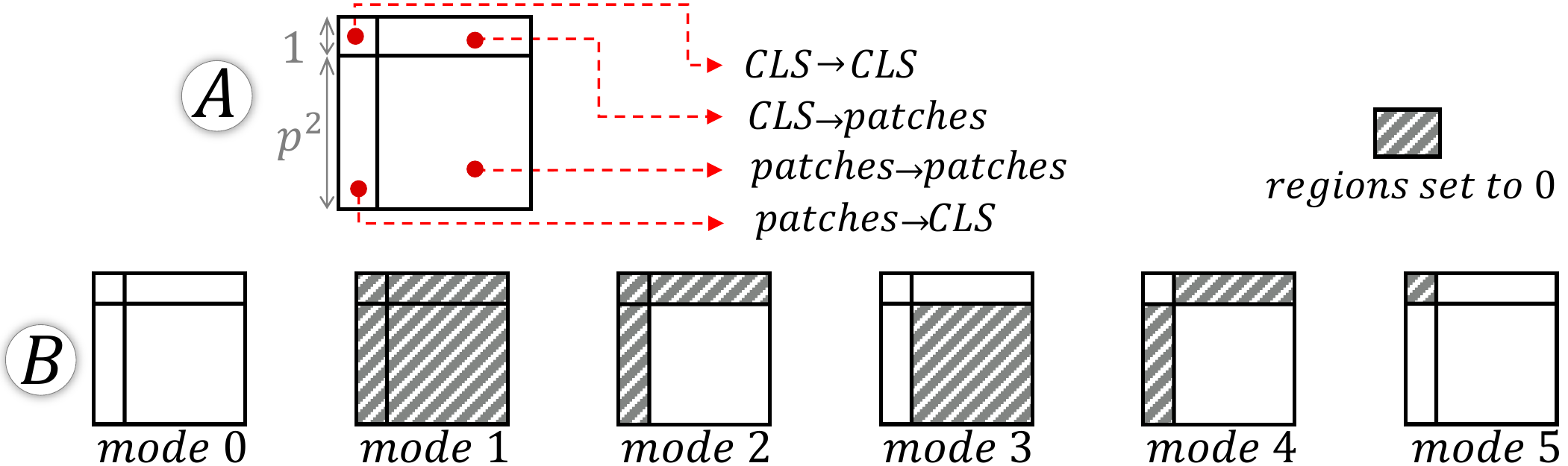}
    \vspace{-0.1in}
    \caption{Pruning modes. (A) The attention matrix is divided into four regions. (B) Different pruning modes set one/multiple regions to 0.}
    \label{fig:pruning_modes}
\end{figure}

As shown in Fig.~\ref{fig:pruning_modes}A, the attention matrix of a head can be divided into four regions based on the source and target tokens: $\texttt{CLS} {\to} \texttt{CLS}$, $\texttt{CLS} {\to} patches$, $patches {\to} \texttt{CLS}$, and $patches {\to} patches$.
Regions $\texttt{CLS}{\to} patches$ and $patches {\to} \texttt{CLS}$ are considered together, as they both encode the interaction between the \texttt{CLS} and patch tokens.
Six pruning modes are defined by setting different regions 
to zero (Fig.~\ref{fig:pruning_modes}B), i.e., mode 0 is the original head without pruning; mode 1 prunes the head completely; modes $2{\sim}5$ are additional cases where only the striped regions are pruned. Showing the impacts from these modes attributes the head's importance to individual regions.

\subsubsection{Visualization}
\label{sec:head_imp_vis}
The \headimportance{} (Fig.~\ref{fig:teaser}B) visualizes our proposed metrics and pruning modes. First, Fig.~\ref{fig:teaser}-B1 uses a line chart to present the four head importance metrics for a single selected image (i.e., the heads' \textit{local} importance to an image). The horizontal axis represents all the $l{\times}n$ heads, and the vertical axis denotes a metric's value, where the dropdown widget enables users to switch among the four metrics. Note that for $I_{i,j}^{prob}$, we directly show the value of $\textbf{P}_{i,j}[idx_{label}]$ (instead of the difference in Eq.~\ref{eq:prob}) as it is more intuitive. When no image is selected (e.g., at the beginning of exploration), the curve in this view shows the average value of the selected metric over all images. Meanwhile, a blue band surrounding the curve denotes the standard deviation of the metric's values (see Fig.~\ref{fig:importance_summary}). The mean and standard deviation reflect the \textit{global} importance over all images, guiding users to select globally important heads. 

Second, after a head is selected from Fig.~\ref{fig:teaser}-B1 (by dragging the vertical line), the bar-chart in Fig.~\ref{fig:teaser}-B2 shows the selected importance metric (y-axis) in different pruning modes (x-axis), further dissecting the head's importance. For example, Fig.~\ref{fig:teaser}-B2 reveals that the importance of the selected head originates from the patch tokens solely, and pruning \texttt{CLS}-related attentions shows no impact.

Lastly, Fig.~\ref{fig:teaser}-B3 shows the top-5 predicted probabilities for the selected image, in the current pruning. If the true label is among the top 5, it will be highlighted in bold.

\subsection{The Attention Strength View}
The attention strength of a head characterizes the spatial distributions of the attention strength across all patches, which answers why the head is important by disclosing where it makes the patches focus.

\subsubsection{Attention Strength Over $k$-Hop Neighbors (\textbf{R2.1})}

We define a $p$-dimensional ($p$D) \textit{attention strength vector}, $\textbf{s}$, for each head, which profiles the average attention strength of all patches to their $k$-hop neighbors ($k\in [0, p{-}1]$), i.e., 
\vspace{-0.1in}
\begin{equation}\small
\label{eq:khop}
    \textbf{s}=\frac{1}{p^2}\Sigma_i\Sigma_j \textbf{s}^{(i,j)},\  i{\in}[0, p{-}1],\ j{\in}[0, p{-}1],
\end{equation}
\vspace{-0.13in}
\begin{equation}
\textbf{s}^{(i,j)} = <s^{(i,j)}_0, s^{(i,j)}_1, s^{(i,j)}_2, \dots, s^{(i,j)}_{p - 1}>,
\end{equation}
where $s^{(i,j)}_k$ denotes the average attention from patch $(i,j)$ to its $k$th hop neighbors in the 2D domain.
\begin{figure}[tbh]
    \centering
    \includegraphics[width=.98\columnwidth]{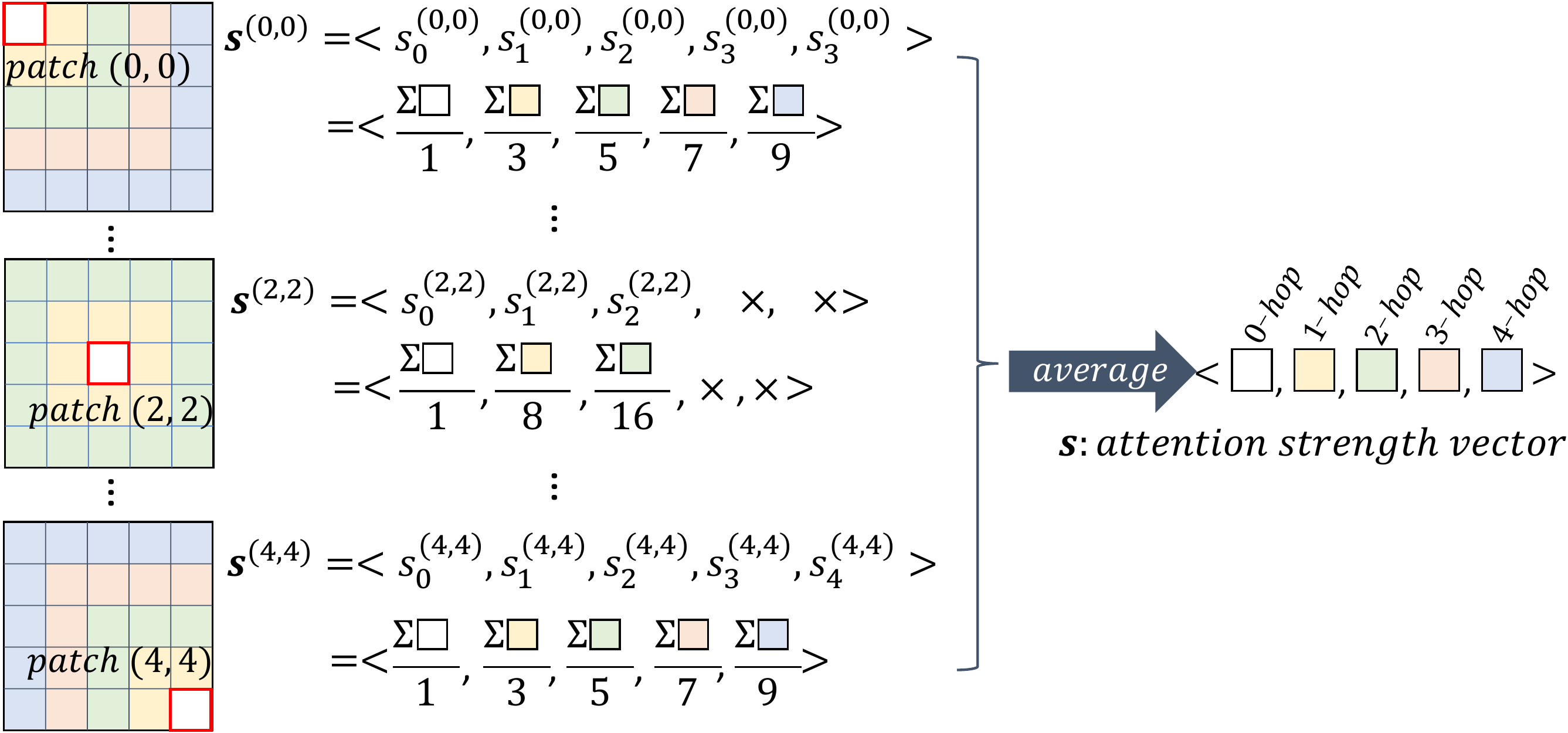}
    \vspace{-0.1in}
    \caption{Computing the attention strength vector for a single head.}
    \label{fig:attention_strengths}
\end{figure}

Without loss of generality, Fig.~\ref{fig:attention_strengths} shows the computation of $\textbf{s}$ when $p{=}5$. Starting from patch $(0,0)$, $s^{(0,0)}_0$ is the attention that patch $(0,0)$ paid to itself (i.e., 0-hop attention); $s^{(0,0)}_1$ is the sum of the attentions paid to its 1-hop neighbors in yellow divided by the number of neighbors (i.e., 3); $s^{(0,0)}_2$ is the total attentions paid to its 2-hop neighbors in green divided by the number of neighbors (i.e., 5); and so on so forth. To the end, we get a $5$D vector for patch (0, 0), i.e., $\textbf{s}^{(0,0)}$. Repeating this computation to all patches, we get $25$ $5$D attention strength vectors, one for each patch. Their average is the head's \textit{attention strength vector}, i.e., $\textbf{s}$.

Note that some patches may not have certain hops of neighbors, e.g., patch $(2,2)$ in Fig.~\ref{fig:attention_strengths} does not have $3$-hop or $4$-hop neighbors (marked as `$\times$'). Therefore, $s^{(2,2)}_3$ and $s^{(2,2)}_4$ will not be counted when computing the corresponding element of vector $\textbf{s}$. In other words, the denominator in Eq.~\ref{eq:khop} is not $p^2$ for all elements of $\textbf{s}$; some will have a smaller denominator due to the missing neighbors.

\subsubsection{Visualization}
The \attnstrength{} (Fig.~\ref{fig:teaser}C) presents all heads' attention strength with three components. The first component (Fig.~\ref{fig:teaser}-C1) presents an overview of all heads for the selected image through a scatterplot. Each point in the scatterplot is a head. Its horizontal position (as well as its color) reflects the layer that the head is from. Its vertical position denotes the entropy of the head's attention strength vector $\textbf{s}$ (normalized). The entropy of $\textbf{s}$ reflects if the head's attention strength is localized on a certain-hop of neighbors (low-entropy, one element's value dominates the vector) or spread across all $k$-hop neighbors (high-entropy, all elements' values are similar). From the overview, there is an obvious trend of the heads across layers (\textbf{R2.2}), i.e., heads from higher layers attend more evenly to all patches, whereas lower-layer heads attend either locally or globally.

Second, after a head is selected by clicking the corresponding point in Fig.~\ref{fig:teaser}-C1, its attention strength vector is presented as a bar chart in Fig.~\ref{fig:teaser}-C2 (\textbf{R2.1}). In the current visualization, we can see that all patches in the selected head attend only to themselves (i.e., all attention strengths are distributed to the 0-hop neighbors).

Lastly, the area plot in Fig.~\ref{fig:teaser}-C3 presents the distribution of entropy values for the selected head over all images (\textbf{R2.3}). For example, the currently selected head has a small entropy (Fig.~\ref{fig:teaser}-C1) as all patches attend to 0-hop neighbors only (Fig.~\ref{fig:teaser}-C2). Fig.~\ref{fig:teaser}-C3 further reveals that the head has consistently low entropy across all images, reflected by the peak on the left corner. The vertical line over the area plot marks the head's entropy for the currently selected image, reflecting how much the head's attention strength for the current image varies from its strength for other images. 

\subsection{The Attention Pattern View}

The attention pattern of a head reflects how tokens are attending to each other. We want to summarize the possible patterns of all heads to deepen the understanding of ViTs. 

\subsubsection{Unsupervised Pattern Identification (\textbf{R3.1})}
\label{sec:pattern_generation}
\begin{figure}[tbh]
    \centering
    \includegraphics[width=\columnwidth]{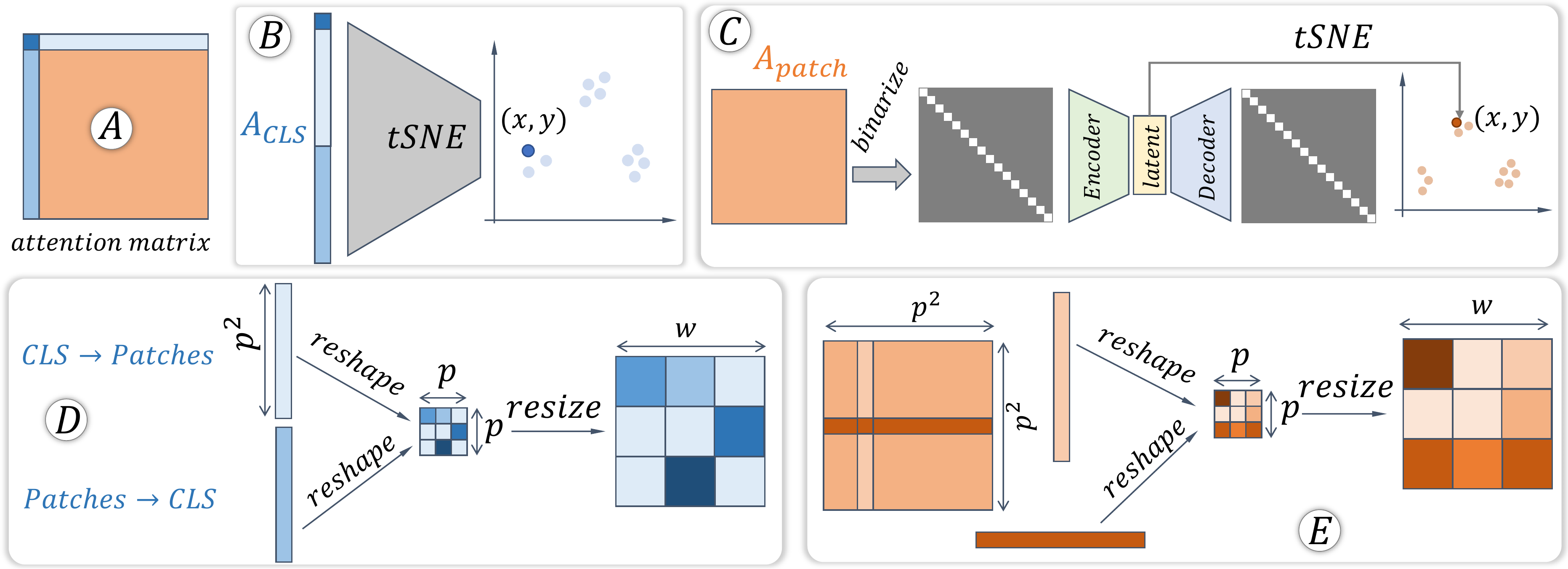}
    \vspace{-0.25in}
    \caption{(A) Each attention matrix is separated into \texttt{CLS}-related attentions ($A_{\texttt{CLS}}$, blue) and patch attentions ($A_{\texttt{patch}}$, orange). (B) $A_{\texttt{CLS}}$ from all images is visualized through tSNE+scatterplot. (C) For $A_{\texttt{patch}}$ from all images, we train an autoencoder to cluster them. (D, E) $A_{\texttt{CLS}}$ and $A_{\texttt{patch}}$ can be mapped back to the image as a mask.}
    \label{fig:attention_pattern_vis}
\end{figure}

Due to the functionality difference between the \texttt{CLS} and patch tokens, we treat them separately and learn their respective patterns.
Specifically, given an input image and one of its heads, the corresponding attention matrix $A$ is of shape ${(1{+}p^2){\times}(1{+}p^2)}$ (Fig.~\ref{fig:architecture}). 
We separate $A$ into \texttt{CLS}-related attentions $A_{\texttt{CLS}} {=} concat(A[0, :], A[1:, 0]) {\in} \mathbb{R}^{2p^2 {+} 1}$ and patch attentions $A_{patch} {=} A[1:, 1:] {\in} \mathbb{R}^{p^2 \times p^2}$, as shown in Fig.~\ref{fig:attention_pattern_vis}A.

\textbf{\texttt{CLS} Attention Patterns.}
The \texttt{CLS}-related attentions ($A_{\texttt{cls}}$) concatenates the $\texttt{CLS}{\to}\texttt{CLS}$, $\texttt{CLS}{\to} patches$, and $patches{\to}\texttt{CLS}$ regions (Fig.~\ref{fig:attention_pattern_vis}A), and its size is $(2p^2{+}1)$. If we have $m$ images, each generates $l{\times}n$ attention matrices from the $l{\times}n$ heads, we will have $m{\times}l{\times}n$ such vectors. Using tSNE, we project them from $(2p^2{+}1)$D to 2D and present them with a scatterplot (Fig.~\ref{fig:attention_pattern_vis}B). Attention heads with similar \texttt{CLS} attention patterns will be clustered together.

\textbf{Patch Attention Patterns.} We applied the same method to the patch attentions $A_{patch} {=} A[1:, 1:] {\in} \mathbb{R}^{p^2 {\times} p^2}$, but the resulting tSNE layout could not clearly separate/cluster dissimilar/similar patch attention patterns. We believe this is caused by the much higher dimensionality of the patch attentions and tried to fix it with several remedies. For example, we used max pooling to spatially shrink $A_{patch}$ before tSNE, and tried to apply PCA on $A_{patch}$ before tSNE. Both solutions did not yield much performance gain.

In the end, we came up with an autoencoder (AE)-based learning solution (Fig.~\ref{fig:attention_pattern_vis}C). \textit{First}, as we care more about the attention pattern, instead of the magnitude, we binarize $A_{patch}$ using a cutoff, e.g., setting top 1\% values to 1 and the rest to 0. This enhances the patterns and makes them easier to learn. \textit{Second}, using the $m{\times}l{\times}n$ binarized $A_{patch}$, we train an AE. The AE has two symmetric subnetworks, i.e., the encoder and decoder, each with two convolutional layers and one fully-connected layer. \textit{Third}, using the latent representations from the well-trained AE's bottleneck layer, we conduct tSNE layout. The layout shows obvious clusters, exposing different attention patterns.

\subsubsection{Visualization}
The \attnpattern{} adopts the ``overview+details'' exploration strategy to visualize the attention patterns.

The \textbf{overview} presents all heads from all images ($m{\times}l{\times}n$ in total) through tSNE+scatterplot (\textbf{R3.1}, Fig.~\ref{fig:teaser}-D1), as explained in Sec.~\ref{sec:pattern_generation}. 
The tSNE layout could be either for the \texttt{CLS} attentions ($A_{\texttt{CLS}}$) or for the patch attentions ($A_{patch}$). The top-right toggle enables this switch.
To be scalable, we allow users to convert the scatterplot into a density plot, and the top-left toggle controls this. For example, the background density contours in Fig.~\ref{fig:teaser}-D1 present the distribution of all the $m{\times}l{\times}n$ heads, as a context. When an image of interest is selected from the \imageoverview{} (Fig.~\ref{fig:teaser}A), its $l{\times}n$ heads will be shown on top of the density plot as points, the color of each reflects its layer.

The \textbf{details} of the attention matrix for a selected head are shown in the right of Fig.~\ref{fig:teaser}D (\textbf{R3.2}). 
An attention matrix denotes the attention between ($1{+}p^2$) tokens, and we present it in two different manners (Fig.~\ref{fig:teaser}-D3, D4).

Fig.~\ref{fig:teaser}-D3 lists all ($1{+}p^2$) tokens as two rows (top: source tokens, bottom: target tokens) and uses lines with light to dark color to encode the attention magnitude. Blue and orange are used to color $A_{\texttt{CLS}}$ and $A_{patch}$ respectively. Showing all the $(1{+}p^2){\times}(1{+}p^2)$ lines would make the view very cluttered. Therefore, we enable users to specify a threshold, the lines with associated attention value below which will be disabled. The histogram in Fig.~\ref{fig:teaser}-D5 shows the distribution of the $(1{+}p^2){\times}(1{+}p^2)$ values, guiding users to specify the threshold by dragging the vertical bar on top of the histogram. The current threshold in Fig.~\ref{fig:teaser}-D5 is 1\%, indicating only the top 1\% lines are visible. 
From the dark vertical lines in Fig.~\ref{fig:teaser}-D3, we can easily see that all tokens (both \texttt{CLS} and patch tokens) strongly attend to themselves.

Fig.~\ref{fig:teaser}-D4 presents the attention matrix through a heatmap (row: source token; column: target token). The four parts of the heatmap have been illustrated in Fig.~\ref{fig:pruning_modes}A. For $A_{patch}$ in the bottom-right corner, one pixel represents one attention value. For $A_{\texttt{CLS}}$, as one pixel is barely visible for the single row and column of attention values, we augment them to take 10 pixels. The color mapping is consistent with that in Fig.~\ref{fig:teaser}-D3 (blue: $A_{\texttt{CLS}}$; orange: $A_{patch}$). 
From the heatmap, we can observe a clear diagonal pattern, indicating that the patch tokens attend strongly to themselves. The \texttt{CLS} token also strongly attends to itself, as the top-left cell is in dark blue. Meanwhile, \texttt{CLS} also attends to different patch tokens, but the attention magnitude is very small (light blue or white color in the top row). The threshold specified from the histogram in Fig.~\ref{fig:teaser}-D5 also applies to this heatmap.

The reason for presenting the attention matrix in two visualizations is to leverage their respective advantages.
The heatmap shows the attention patterns more intuitively, whereas the two-axes view can better present the attention relationship between the \texttt{CLS} and patch tokens (one example is shown later in Fig.~\ref{fig:bottom}). The two-axes view is also better than the heatmap in terms of interacting with tokens, e.g., it can easily highlight a token of interest (see Fig.~\ref{fig:case_head137}).
Apart from these two, we have also considered other visualizations in our early design stages. For example, we tried to overlay arrows on top of the heatmap to show the attention direction or embed patch pixels into the heatmap. However, both designs are not easily scalable to our problem size.

\textbf{Image Context.} To intuitively present the patch-related attentions, we need to map the patches back onto the 2D image. For $\texttt{CLS}{\to}patches$ (shape: $1{\times}p^2$) and $patches{\to}\texttt{CLS}$ (shape: $p^2{\times}1$) attentions, we reshape them to a $p{\times}p$ square, scale the square to $w{\times}w$, and overlay it on top of the image as a transparency mask (Fig.~\ref{fig:attention_pattern_vis}D, stronger attention${\to}$more transparent). For the $patches{\to}patches$ attentions (shape: $p^2{\times}p^2$), we reshape individual row/column into a $p{\times}p$ square, scale it to $w{\times}w$, and overlay it on top of the image as a mask to show the attention from a token to all others (a row) or vise versa (a column), Fig.~\ref{fig:attention_pattern_vis}E.
The three images in Fig.~\ref{fig:teaser}-D2 (top-bottom) show the original image, image+source attention mask, and image+target attention mask. 
Hovering over individual source/target tokens from the top/bottom axis in Fig.~\ref{fig:teaser}-D3 will update the source/target attention masks dynamically (e.g., Fig.~\ref{fig:case_head10_cls}D,  Fig.~\ref{fig:case_head137}C).

\section{Case Study and Experts' Feedback}
\label{sec:case_study}

We use multiple case studies, conducted together with deep learning experts, to show the capability of our system. The experts' feedback is summarized at the end of Sec.~\ref{sec:feedback}.

For ViTs, we explored four pre-trained ViTs with different image resolutions ($w$), numbers of layers ($l$) and heads ($n$)~\cite{dosovitskiy2020image}. As our findings are consistent across them, we focus our illustrations on one model only but include results of the other three in our Appendix. The parameters of the focused model are: $w{=}224$, $l{=}12$, $n{=}12$, and $p{=}14$.

For datasets, we used 1000 images sampled from the validation set of ImageNet~\cite{russakovsky2015imagenet}. The images are from 20 classes (10 classes with the best and 10 classes with the worst predictions), each has 50 images. 
We have also explored our system with another dataset. However, as the model-level findings from the two datasets are mostly similar, we include the exploration of the other dataset in our Appendix.

\subsection{Head Importance}
\label{sec:case_head_importance}
Fig.~\ref{fig:teaser}A shows the 1000 images, laid out by their $l{\times}n$-dimensional head importance vector (each dimension is $\textbf{P}_{i,j}[idx_{label}]$ in Eq.~\ref{eq:prob}). From the layout, we found images are arranged clockwise with an increasing true-class probability. 

\begin{figure}[tbh]
    \centering
    \includegraphics[width=\columnwidth]{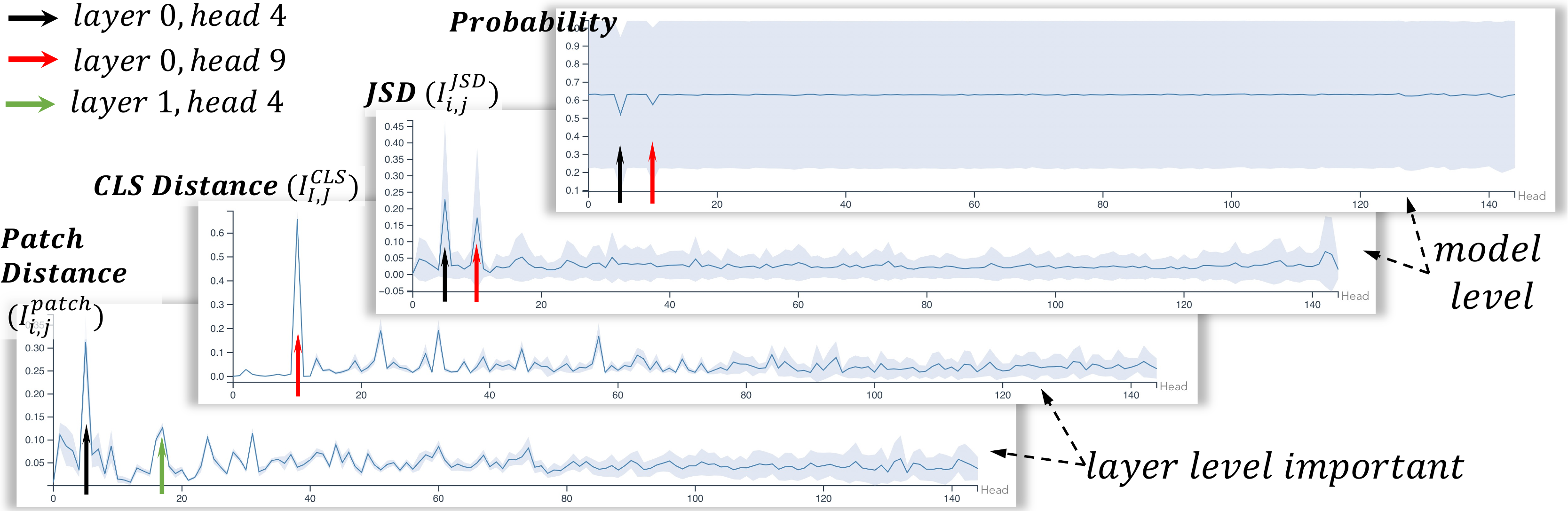}
    \vspace{-0.2in}
    \caption{The mean (blue curve) and standard deviation (blue band) of the four head importance metrics computed over all images.}
    \label{fig:importance_summary}
\end{figure}

Fig.~\ref{fig:importance_summary} shows the four head importance metrics aggregated over all images. Both \textit{model-level} metrics (`Probability', `JSD') reflect head $4$ and $9$ from layer 0 are very important. The very wide band of the `Probability' plot is due to our choice of images with the best and worst performances.
The two \textit{layer-level} metrics (`\texttt{CLS} Distance', `Patch Distance') show heads 9 and 4 contribute significantly to the changes of \texttt{CLS} and patch representations, respectively.

Moreover, the \textit{layer-level} metrics show more oscillations, identifying important heads that cannot be identified from the \textit{model-level} metrics.
For example, removing layer 1 head 4 barely changes the final predictions but significantly affects the layer activations. As shown in the `Patch Distance' plot, the mean for this head is large and its standard deviation is small, 
indicating the head is important to the layer across images.
This head has a fixed function of making all patches attend to the patch above themselves (explained later in Fig.~\ref{fig:tsne_pattern}F). The \textit{model-level} metrics cannot identify it as heads from later layers show similar functionalities (i.e., head 3 from layer 2, explained later in Fig.~\ref{fig:tsne_pattern}J), hiding its importance. Also, an increasing standard deviation is observed from the two \textit{layer-level} metrics, indicating that higher-layer heads' importance is more influenced by image contents.

By coordinately exploring the \imageoverview{} and \headimportance{}, we find image clusters, to which, individual heads are very important (e.g., heads 9 and 4 from layer 0 are important to images in Fig.~\ref{fig:teaser}-A1, A2). To analyze their importance, we randomly select an image (ID: 712, label: \texttt{macaw}) from one cluster for further exploration. The three heads that are very important to the selected image are: \headfour{}, \headnine{}, and \headfive{} (Fig.~\ref{fig:teaser}-B1). 

Fig.~\ref{fig:teaser}-B2 shows the partial pruning results for \headfour{}. The probability drops only if the $patches{\to}patches$ attentions 
are pruned, and pruning \texttt{CLS}-related attentions has little impact. This attributes head 4's importance to its attentions between patches.  Fig.~\ref{fig:teaser}-B3 shows the top-five probabilities are very low when this head is pruned, and \texttt{macaw} (the true label) is not among them.

Fig.~\ref{fig:teaser}-B4 and B5 show the partial pruning results for \headnine{} and \headfive{}, respectively. From them, \headnine{} is important due to the \texttt{CLS} self-attention ($\texttt{CLS}{\to}\texttt{CLS}$); \headfive{} is important due to the attentions between \texttt{CLS} and patches ($\texttt{CLS}{\to}patches$ and $patches{\to}\texttt{CLS}$).

\subsection{Attention Strength}
\label{sec:case_strength}
Next, we examine why the heads are important through their attention strengths. The attention strength analysis is for patch tokens only (\texttt{CLS} has no spatial information), so our analysis focuses on \headfour{}.
Fig.~\ref{fig:teaser}-C1 shows the attention strength overview of all heads. \headfour{} (in the red circle) has the smallest entropy. Fig.~\ref{fig:teaser}-C2 shows the head's $p$D $k$-hop neighborhood vector. From it, all patches' attention strengths focus on the $0$-hop neighbor, i.e., the head makes all patches strongly attend to themselves. It is reasonable that such a head is important in lower layers, as no details should be overlooked at the beginning. From Fig.~\ref{fig:teaser}-C3, we also notice that this head's functionality is consistent across all images, as its entropies for different images are always low (distributed dominantly to the left).

Fig.~\ref{fig:teaser}-C4, C5 show the attention strengths of the other two important heads. Their importance majorly comes from \texttt{CLS}-related attentions, and the attention strengths (for patch tokens) are scattered across all hops of neighbors (the bar chart) and varying across images (the area plot).
\begin{figure}[tbh]
    \centering
    \includegraphics[width=0.97\columnwidth]{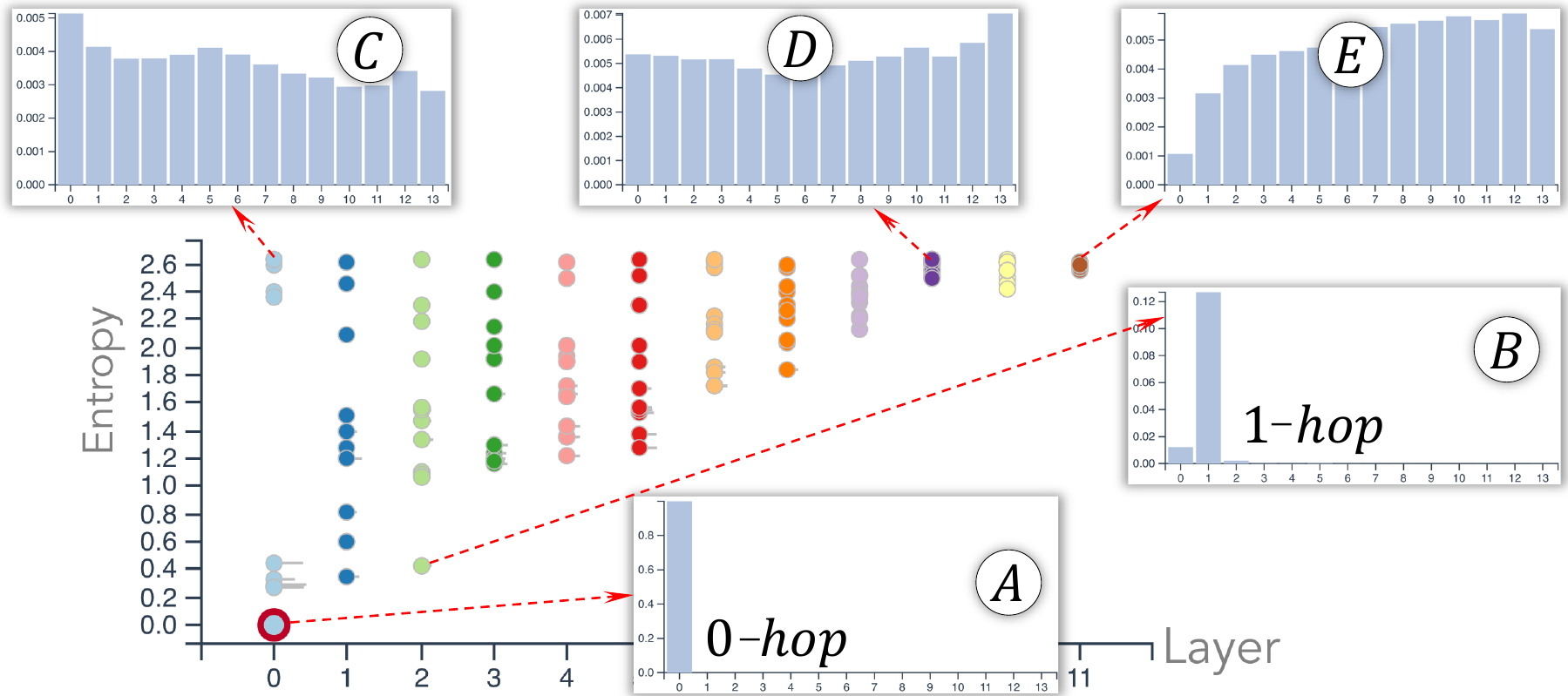}
    \vspace{-0.1in}
    \caption{Lower-layer heads can distribute patches' attention strength only to their near neighbors (A, B), or evenly to all $k$-hop neighbors (C). Higher-layer heads can only do the latter (D, E).}
    \label{fig:attention_strength}
\end{figure}

The overview in Fig.~\ref{fig:teaser}-C1 and Fig.~\ref{fig:attention_strength} also reveals the attention strength distribution of heads over layers. 
In general, lower-layer heads can make patches attend strongly to their local regions, e.g., $0$-hop or $1$-hop neighbors (the low entropy heads in Fig.~\ref{fig:attention_strength}A-B). Low-layer heads can also make patches evenly distribute their attention across the entire image (e.g., Fig.~\ref{fig:attention_strength}C). For higher-layer heads, patches only attend globally with similar attention strengths across all $k$-hop neighbors, e.g., Fig.~\ref{fig:attention_strength}D-E. 
A similar overview can always be observed no matter which image is selected.
This observation is consistent with the original claims about attention strengths~\cite{dosovitskiy2020image}
(see details about~\cite{dosovitskiy2020image} in our Appendix).

\subsection{Attention Pattern}
\label{sec:case_pattern}

The attention patterns further answer why \headfour{} is important. From the \attnpattern{}, i.e., the vertical lines in Fig.~\ref{fig:teaser}-D3 and the diagonal pattern in Fig.~\ref{fig:teaser}-D4, all patches in this head strongly attend to themselves, echoing our earlier findings from the \attnstrength{}. 

\begin{figure}[tbh]
    \centering
    \includegraphics[width=\columnwidth]{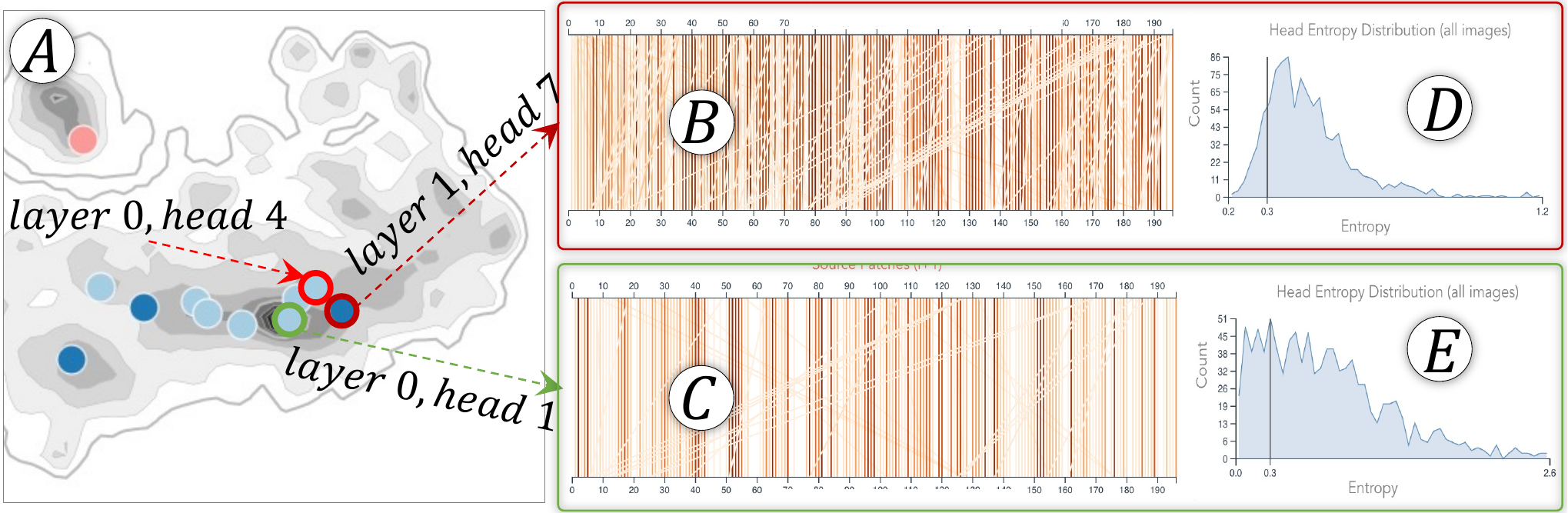}
    \vspace{-0.3in}
    \caption{Exploring heads that have similar attention patterns with \headfour{} to explain why it is important while others are not.}
    \label{fig:case_head5_patterns}
\end{figure}

Are there heads with attention patterns similar to {\textit{head 4}}? If yes, why is only {\textit{head 4}} so important? To answer these questions, we zoom into the red dashed region of the tSNE layout in Fig.~\ref{fig:teaser}-D1. The zoomed-in details are shown in Fig.~\ref{fig:case_head5_patterns}A. From it, we explore heads close to {\textit{head 4}} and check their attention patterns. Fig.~\ref{fig:case_head5_patterns}B-C show two of them, where the vertical lines indicate the patches also majorly attend to themselves in these two heads. However, different from {\textit{head 4}}, the self-attentions in these two heads are not always strong, i.e., all lines in Fig.~\ref{fig:teaser}-D3 are in dark orange, but most lines in Fig.~\ref{fig:case_head5_patterns}B-C are in light orange.
Moreover, the two heads' functionality is not as consistent as that of {\textit{head 4}} across images (comparing Fig.~\ref{fig:teaser}-C3 and Fig.~\ref{fig:case_head5_patterns}D-E).

\begin{figure}[tbh]
    \centering
    \includegraphics[width=\columnwidth]{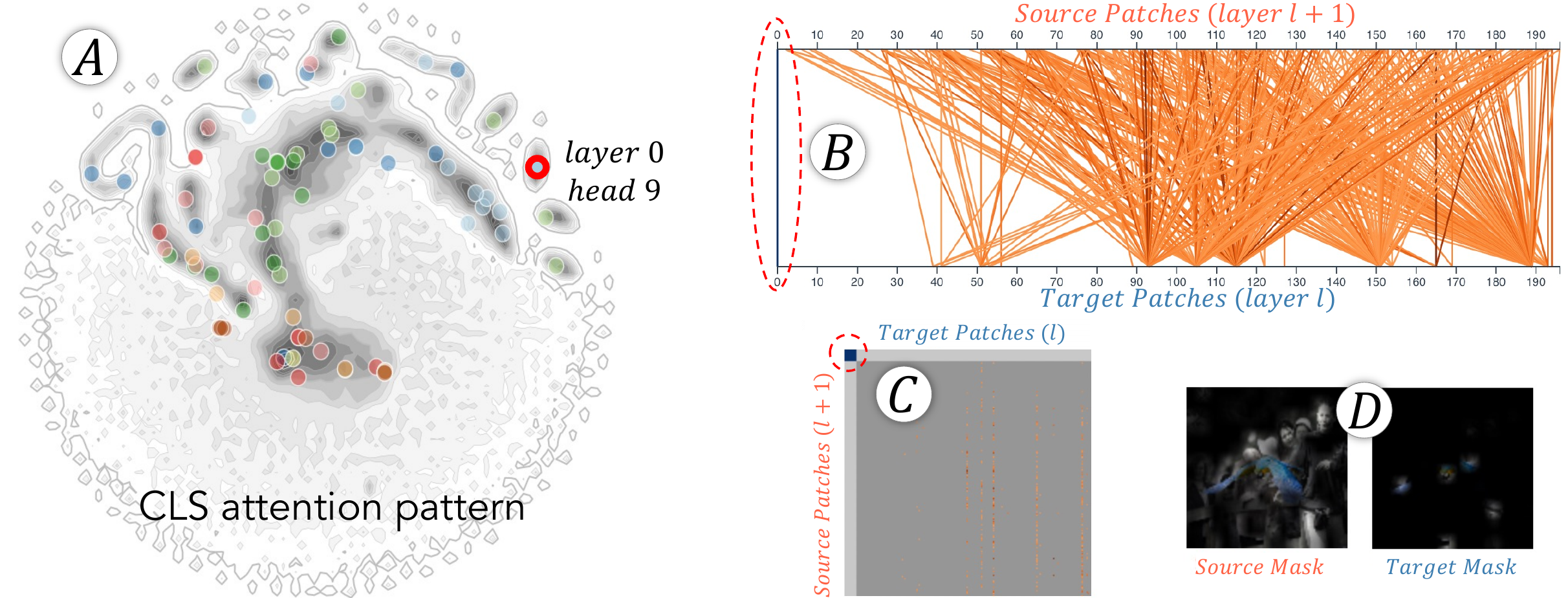}
    \vspace{-0.2in}
    \caption{The attention pattern of \headnine{}.}
    \label{fig:case_head10_cls}
\end{figure}

\begin{figure}[b]
    \centering
    \includegraphics[width=\columnwidth]{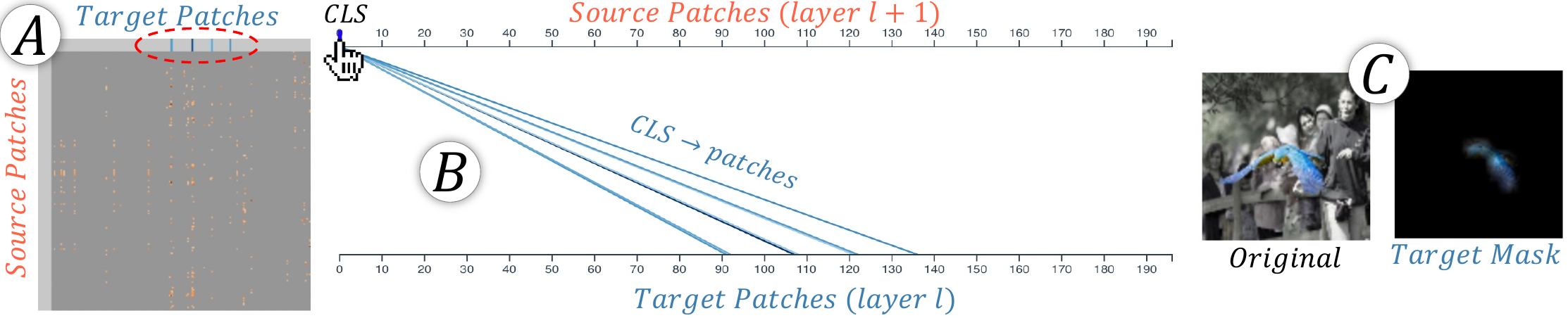}
    \vspace{-0.25in}
    \caption{The $\texttt{CLS}{\to}patches$ attention in \headfive{}.}
    \label{fig:case_head137}
\end{figure}

\begin{figure*}[tbh]
    \centering
    \includegraphics[width=\textwidth]{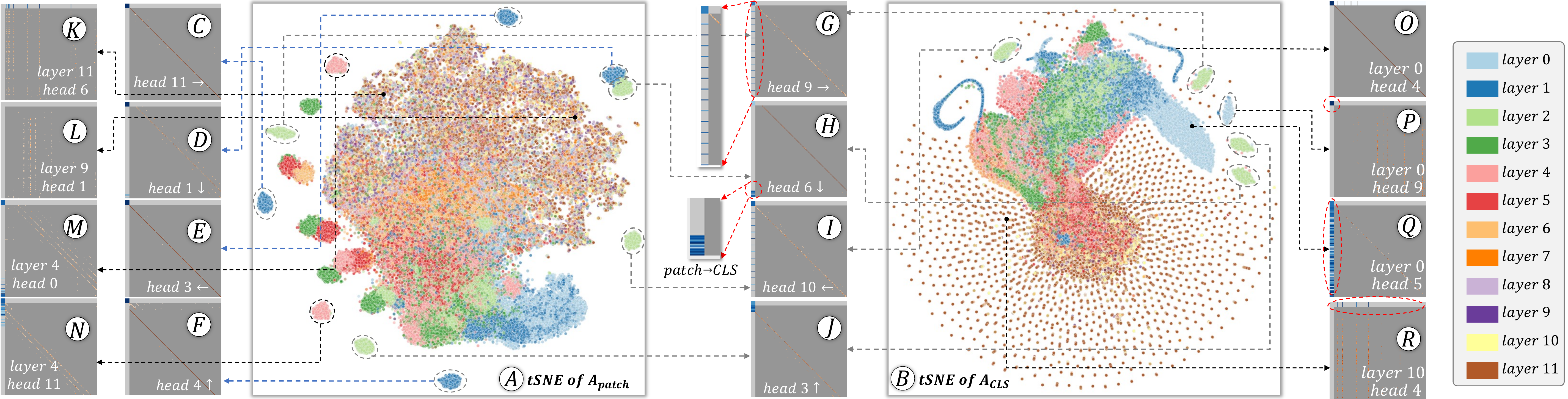}
    \vspace{-0.3in}
    \caption{(A/B) tSNE layouts of heads using their patch/\texttt{CLS} attention patterns. Content-agnostic heads from lower layers show similar patterns across images, and thus each forms an isolated cluster no matter they are laid out by the patch (C-J, M, N) or \texttt{CLS} (G-J, O-Q) patterns. Content-relevant heads from higher layers show dissimilar patterns for different images. These heads are scattered in both layouts (K, L, R).
    }
    \label{fig:tsne_pattern}
\end{figure*}

\begin{figure*}[tbh]
 \centering 
 \includegraphics[width=\textwidth]{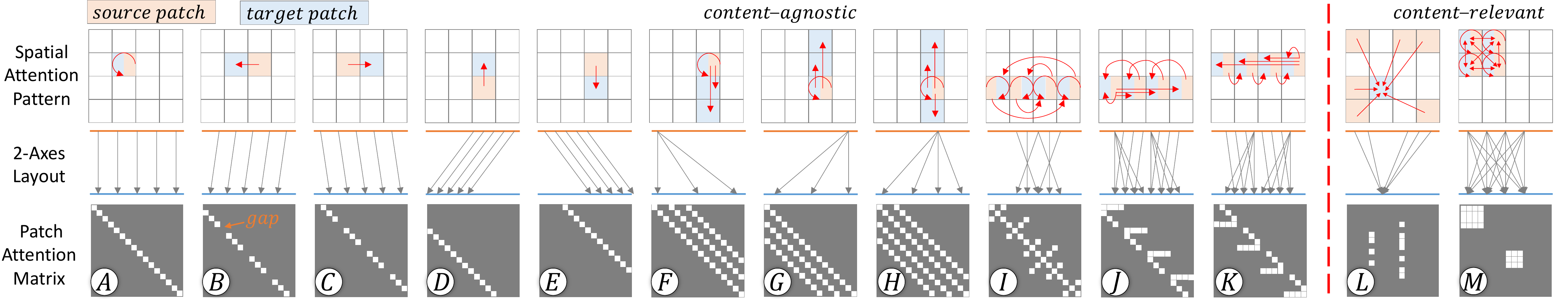}
  \vspace{-0.3in}
 \caption{The 13 possible attention patterns between image patches presented in the (top) image space, (middle) two-axes, and (bottom) heatmap visualization. Each pattern is a mix of one/multiple of the four basic patterns: \textit{diagonal} (A-K), \textit{horizontal} (J, K), \textit{vertical} (L), and \textit{block} (M). The first two are \textit{content-agnostic} and often appear in lower-layer heads; the latter two are \textit{content-relevant} and often occur in higher-layer heads. }
 \label{fig:pattern}
 \vspace{-0.1in}
\end{figure*}

For \headnine{}, we have known its importance comes from the $\texttt{CLS}{\to}\texttt{CLS}$ attention in Fig.~\ref{fig:teaser}-B4. Visualizing its attention patterns, we can see the dark blue vertical line in Fig.~\ref{fig:case_head10_cls}B and the dark blue cell in the top-left corner of Fig.~\ref{fig:case_head10_cls}C, echoing the importance of \texttt{CLS}'s self-attention. Meanwhile, the $\texttt{CLS}{\to}patches$ and $patches{\to}\texttt{CLS}$ attentions are not noticeable. The $patches{\to}patches$ attentions (the orange lines in Fig.~\ref{fig:case_head10_cls}B and the vertical pattern in Fig.~\ref{fig:case_head10_cls}C), as well as the masked source and target images (Fig.~\ref{fig:case_head10_cls}D), show no obvious extracted features, confirming the less importance of the patch-related attentions. Also, we noticed that the \texttt{CLS} attention pattern of {\textit{head 9}} is very unique, as it is the only head (of the selected image) in the isolated cluster of Fig.~\ref{fig:case_head10_cls}A (the tSNE layout of all heads' $A_\texttt{CLS}$). The background contour shows the head 9 from other images, reflecting that head 9's functionality is fixed across images. 

For \headfive{}, its importance is from the $patches$${\to}$ $\texttt{CLS}$ and
$\texttt{CLS}{\to}patches$ attentions (Fig.~\ref{fig:teaser}-B5). In Fig.~\ref{fig:case_head137}A, the $patches{\to}\texttt{CLS}$ attentions (the leftmost column) are not noticeable, but the $\texttt{CLS}{\to}patches$ attentions (the top row) show four major regions, indicating the \texttt{CLS} attends to four groups of patches. Hovering over the \texttt{CLS} token on the top axis of Fig.~\ref{fig:case_head137}B, the four groups of target patches are highlighted in the masked target image (Fig.~\ref{fig:case_head137}C). From it, the four regions accurately extract the \texttt{macaw}'s wings in blue.

\subsubsection{Attention Pattern Summary}
\label{sec:attn_pattern_summary}
Our explorations identified different attention patterns. This section provides an exhaustive summary.

\textbf{Patch Attention ($A_{Patch}$) Patterns.}
Fig.~\ref{fig:tsne_pattern}A shows the tSNE layout of all heads' patch attentions from the 1000 images (Fig.~\ref{fig:teaser}-D1 is the density plot of it). 
The points from an isolated cluster often represent the same head from different images, verifying the head's fixed functionality.
Exploring individual heads in the \attnpattern{}, we summarize them into 13 patterns in Fig.~\ref{fig:pattern}. The three rows of the illustrative figure show how each patch attends to others in the image space (top), the two-axes (middle), and the heatmap (bottom). From the last row, each of the 13 patterns is a combination of four basic patterns, i.e., \textit{\textbf{diagonal}}, \textit{\textbf{horizontal}}, \textit{\textbf{vertical}}, and \textit{\textbf{block}}.

Fig.~\ref{fig:pattern}A-K include the \textit{\textbf{diagonal} pattern}. Fig.~\ref{fig:pattern}A denotes self-attention. Fig.~\ref{fig:pattern}B-C show each patch attends to its left or right patch (one cell off the diagonal), where the gaps indicate the leftmost or rightmost patches without further ones to attend to. Each patch in Fig.~\ref{fig:pattern}D-E attends to the patch above or below itself, i.e., the white-squares shift from the diagonal for a row of patches.
The four heads in Fig.~\ref{fig:tsne_pattern}C-F (all from layer 1) show examples where each patch attends to its right, bottom, left, and top patch, respectively.
The four heads in Fig.~\ref{fig:tsne_pattern}G-J (all from layer 2) show similar patterns, revealing the repeating functionalities.
Coming to middle attention layers, patches can attend to multiple patches above and/or below themselves (Fig.~\ref{fig:pattern}F-H). Fig.~\ref{fig:tsne_pattern}M-N show two such examples from layer 4. Fig.~\ref{fig:pattern}I includes the counter-diagonal patterns, in which the left/right patches symmetrically attend to the right/left ones in the same row. 

Fig.~\ref{fig:pattern}J-K contain the \textit{\textbf{horizontal} pattern} (mixed with the diagonal pattern). The pattern indicates that a patch attends to multiple patches before/after itself in the same row. 

Fig.~\ref{fig:pattern}L shows the \textit{\textbf{vertical} pattern}, i.e.,  multiple source patches (heatmap rows) attend to the same target patches (heatmap columns). This usually indicates the target patches include important semantics to the class (e.g., the \texttt{cat} face region of a \texttt{cat} image, Fig.~\ref{fig:tsne_pattern}K-L). The pattern often occurs in higher-layer heads, as indicated by the yellow/brown color in Fig.~\ref{fig:tsne_pattern}A (the big chaotic cluster in the center).

Fig.~\ref{fig:pattern}M shows the \textit{\textbf{block} pattern}, i.e., patches in a local region mutually attend to each other (e.g., the face patches of a \texttt{cat} attend to its body patches and vice versa). 
Similar to segmentation, the attentions extract the object's pixels.

Attention patterns can also be categorized into \textit{\textbf{content-agnostic}} and \textit{\textbf{content-relevant}}. The \textit{diagonal} and \textit{horizontal} patterns are often agnostic to the image content, e.g., heads in Fig.~\ref{fig:tsne_pattern}C-J. The same head from all images forms an isolated cluster in Fig.~\ref{fig:tsne_pattern}A.
The \textit{vertical} and \textit{block} patterns are content-relevant, as their position depends on the content of images. They are the big chaotic cluster in Fig.~\ref{fig:tsne_pattern}A.
Content-agnostic patterns often occur in lower layers, whereas content-relevant patterns often occur in higher layers.

\begin{figure}[tb]
    \centering
    \includegraphics[width=0.95\columnwidth]{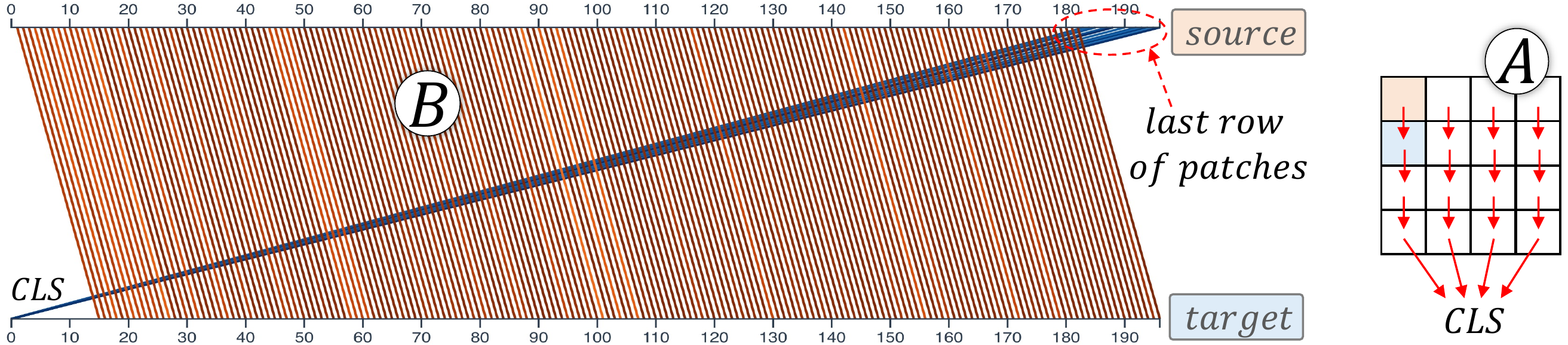}
    \vspace{-0.12in}
    \caption{Each patch attends to the one beneath it, e.g., patch 0 (row 0, column 0) attends to patch 14 (row 1, column 0,  $p{=}14$). The bottom row of patches all attend to \texttt{CLS} since there is no further patch.}
    \label{fig:bottom}
\end{figure}

\textbf{\texttt{CLS} Attention ($A_{\texttt{CLS}}$) Patterns.}
The \texttt{CLS}-related attention patterns follow a similar layer-wise trend. 
As shown in Fig.~\ref{fig:tsne_pattern}B (its density plot is in Fig.~\ref{fig:case_head10_cls}A), heads in lower layers form clear clusters, indicating the \texttt{CLS}'s attentions in them are more content-agnostic.
For example, Fig.~\ref{fig:tsne_pattern}P shows dominantly strong $\texttt{CLS}{\to}\texttt{CLS}$ attentions; Fig.~\ref{fig:tsne_pattern}Q shows strong $patches{\to}\texttt{CLS}$ attentions.
In higher layers, the attentions are more content-relevant, e.g., in Fig.~\ref{fig:tsne_pattern}R (the top row), the \texttt{CLS} focuses only on specific image patches.

The four heads with isolated clusters in Fig.~\ref{fig:tsne_pattern}A, G-J also form four isolated clusters in their \texttt{CLS} layout (Fig.~\ref{fig:tsne_pattern}B, G-J). By coordinately exploring the patch and \texttt{CLS} patterns, we found the boundary patches without further patch to attend in these heads will attend to \texttt{CLS}. For example, in Fig.~\ref{fig:tsne_pattern}H, all patches attend to the patch beneath themselves. The bottom row of patches have no patches beneath them, so they attend to \texttt{CLS} (Fig.~\ref{fig:bottom}A). The two-axes view of this head is shown in Fig.~\ref{fig:bottom}B, which is consistent with the pattern in Fig.~\ref{fig:tsne_pattern}H (and the inset). 
This explains how information is passed across patches row-by-row to \texttt{CLS} for classification.

\subsection{Head Attention Diagnosis}
Our coordinated system also helps to diagnose the roles of different heads (especially the important ones) in mispredictions. We brief two example cases in this section.

\begin{figure}[tbh]
    \centering
    \includegraphics[width=\columnwidth]{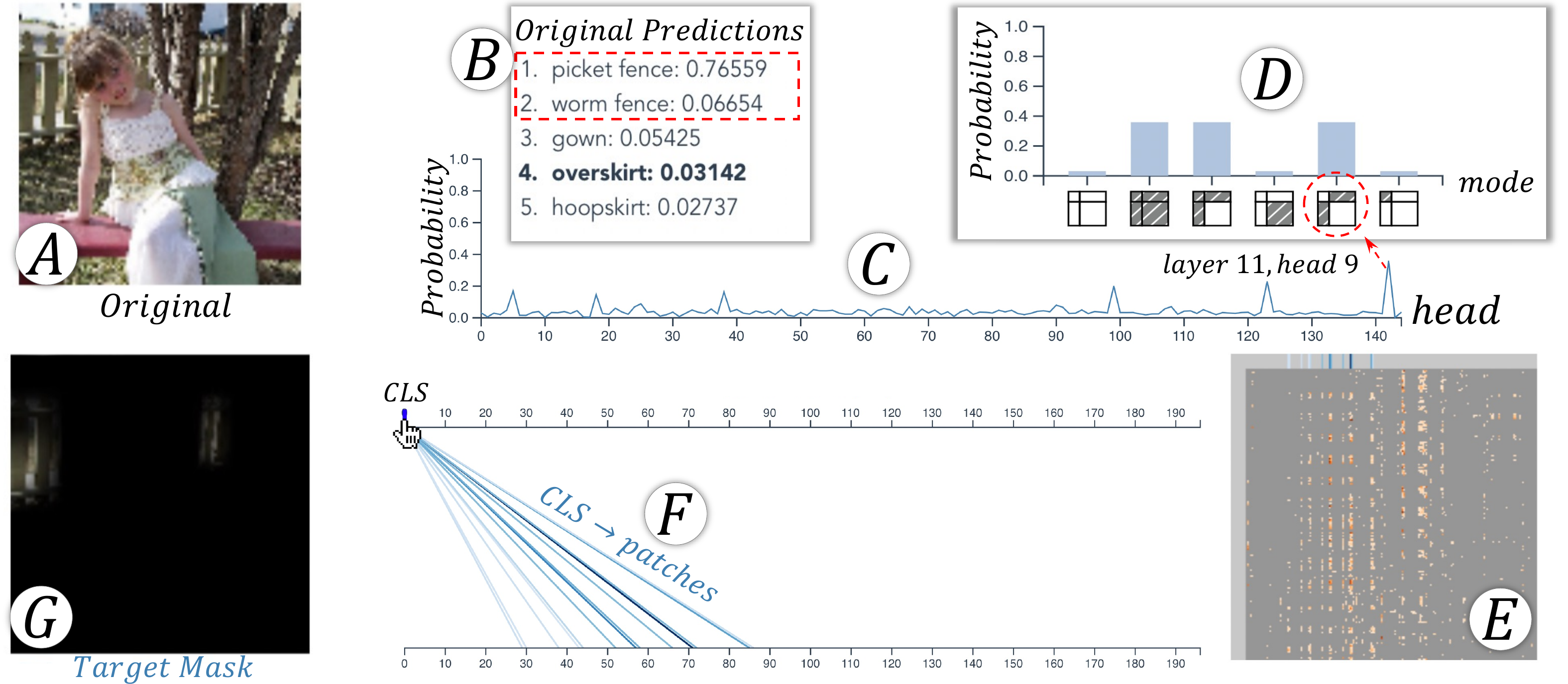}
    \vspace{-0.3in}
    \caption{The head 9 from layer 11 incorrectly attends to the background fence. Pruning it will increase the true class's probability.}
    \label{fig:case_image355}
\end{figure}

\textit{\textbf{Case 1:}} Fig.~\ref{fig:case_image355}A shows an image from the \texttt{overskirt} class, but the ViT performs badly on its prediction (Fig.~\ref{fig:case_image355}B). From the \headimportance{}, we notice a sharp increase in the true label probability when head 9 of layer 11 is pruned (Fig.~\ref{fig:case_image355}C). The partial pruning result of this head (Fig.~\ref{fig:case_image355}D) reflects that the probability of \texttt{overskirt} will \textit{increase} only if the attentions between \texttt{CLS} and patches are pruned.
As the $patches{\to}\texttt{CLS}$ attentions are very small in Fig.~\ref{fig:case_image355}E (the first column), we hover over the $\texttt{CLS}{\to}patches$ attentions in Fig.~\ref{fig:case_image355}F. The masked target image (Fig.~\ref{fig:case_image355}G) shows that the \texttt{CLS} attends strongly to the background fences. Pruning such a misleading head will help the model focus more on the right features, leading to a probability increase. The importance of this head also explains why the top two classes in the original predictions (Fig.~\ref{fig:case_image355}B) are \texttt{fences}.

\begin{figure}[tbh]
    \centering
    \includegraphics[width=\columnwidth]{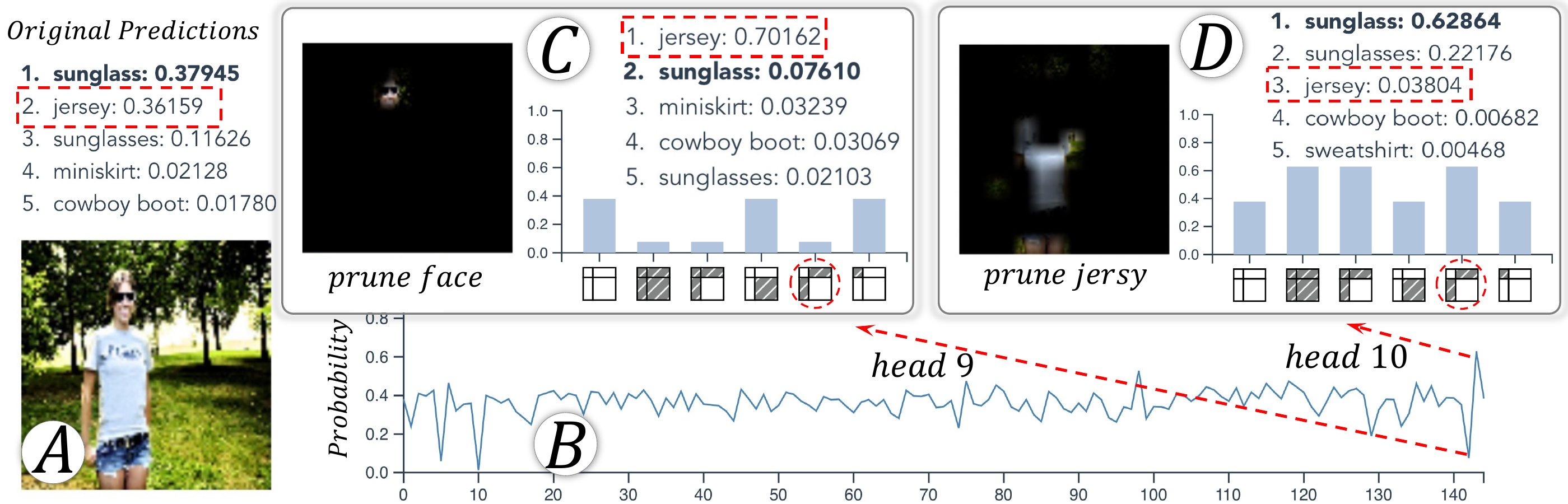}
    \vspace{-0.3in}
    \caption{Heads 9 and 10 (layer 11) extract the \texttt{sunglass} and \texttt{jersey} features respectively, impacting the corresponding classes' probability.}
    \label{fig:case_image483}
\end{figure}

\textit{\textbf{Case 2:}} The ViT correctly predicts the image in Fig.~\ref{fig:case_image483}A to be \texttt{sunglass}, but also assigns a high probability to \texttt{jersey}. 
We found two important heads in the last layer but with opposite effects (Fig.~\ref{fig:case_image483}B). Pruning heads 9 and 10 from layer 11 separately leads to a big decrease and increase of the \texttt{sunglass} probability.
From the partial pruning results, we found the attentions between \texttt{CLS} and patches dominate the decrease (Fig.~\ref{fig:case_image483}C) or increase (Fig.~\ref{fig:case_image483}D). We then visualize the $\texttt{CLS}{\to}patches$ attentions and the corresponding masked images. 
For head 9, the \texttt{CLS} focuses on the face region, whereas for head 10, the \texttt{CLS} attends solely to the jersey region. Heads 9 and 10 contribute largely to the probability of \texttt{sunglass} and \texttt{jersey} respectively. Pruning them increases the opposite class's probability (Fig.~\ref{fig:case_image483}C, D). These details significantly deepens the understanding of how ViT works.

\vspace{-0.15in}

\subsection{Domain Experts' Feedback}
\label{sec:feedback}
The above case studies were conducted with 7 deep learning experts in separate sessions, using the protocol of \textit{guided exploration+think-aloud discussions}. All experts are researchers with 5+ years of experience in deep learning. Five experts ($E_1{\sim}E_5$) have participated in our requirement analysis. The other two ($E_6{\sim}E_7$) had no knowledge about our visualization system until the case study sessions.

In general, all experts confirmed the importance of the three focused topics and appreciated our findings. $E_2$, $E_4$, and $E_5$ enjoyed the system's interactivity, especially the linked visualizations, which helped them connect the dots for comprehensive interpretations of important heads. Their existing visualization tools with piece-by-piece analysis fall short of such coordinated explorations.
Using our system, $E_1$ and $E_3$ obtained an overview of ViTs' head attention patterns for the first time. Both experts found our findings intriguing and had thorough discussions on the analogy between CNNs and ViTs.
For example, patches in Fig.~\ref{fig:tsne_pattern}G always attend to their right patch. The $3{\times}3$ CNN filter \inlinegraphics{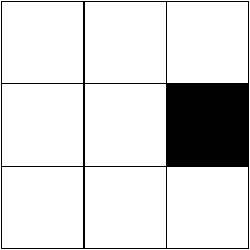} also aggregates the right pixel's value to the current pixel. So, the ViT head and CNN filter share equivalent functions. 
Similar equivalence analysis can also be extended to other heads/filters.
$E_7$ pointed out similar learning trends across layers of CNNs and ViTs, i.e., lower-layer heads/filters focus on local features, whereas higher layers aggregate the output of lower layers to extract object-level information. 
$E_7$ also found the interactions between the \texttt{CLS} and patches very insightful. In Fig.~\ref{fig:bottom}A, all patches attend to the patch below them and the last row attends to the \texttt{CLS}. This is similar to propagating the information top-down, and the final accumulated results are passed to \texttt{CLS} for classification.
$E_6$ was interested in the content-agnostic/relevant attentions and believed they could be adopted for anomaly detection.

The experts also pointed out some insufficiency of the current system. First, $E_7$ initially thought only the three heads in Fig.~\ref{fig:teaser}-B1 were important to the prediction of image 712, which was misleading as we only pruned one head at a time and did not consider the dependency between heads. Second, both $E_1$ and $E_7$ worked on token pruning of transformers (instead of head pruning). They liked our image masks in disclosing the tokens' semantics but also wanted to see similar saliency maps highlighting the importance of individual patches through ablations. These comments provide promising future directions for us to explore.

\section{Discussion, Limitations, and Future Work}

Despite many visual interpretation works for DL, \textit{\textbf{the unique values of our work}} come from the following perspectives. 
First, our work presents a comprehensive interpretation of ViTs and discloses insightful findings.
For example, heads with strong self-attentions
are dominantly important. Lower- and higher-layer heads show different local/global attention strengths. Also, we summarize all possible attention patterns between patches. These insights open the hood of ViTs and deepen model designers' understanding.
Second, our interpretation triggers model improvement ideas, e.g., pruning heads with repeating patterns. Thus, improving ViTs with our derived insights would be a direct follow-up work.
Lastly, although we focus only on the classification task, we believe our interpretations are transferable to ViT-based detection/generation tasks~\cite{khan2021transformers}, as those tasks also significantly rely on the multi-head self-attentions of ViTs.

\textbf{\textit{Head-Centric v.s. Image-Centric.}}
We want to emphasize that all our analyses are \textit{head-centric}, and each head's behavior is analyzed in one and across all images.
Specifically, for \textit{head importance}, we provide each head's \textit{local} importance on one image and \textit{global} importance over all images (Sec.~\ref{sec:head_imp_vis}). For \textit{head attention strength}, we present a head's attention strengths in one image (Fig.~\ref{fig:teaser}-C2) and its strength distribution over all images (Fig.~\ref{fig:teaser}-C3). For \textit{head attention pattern}, the two-axes/heatmap (Fig.~\ref{fig:teaser}-D3, D4) shows the attention pattern of a head from one image, while the scatterplot in Fig.~\ref{fig:teaser}-D1 lays out the head's attention pattern over all images. From a different perspective, we believe \textit{image-centric} analysis would also lead to insightful findings, e.g., checking if the heads show similar patterns for images of the same class. We plan to explore this direction in the future.

\textbf{\textit{Performance.}} 
To guarantee the exploration interactivity, we have pre-computed some of the visualization data.
For example, the head importance metrics are computed offline as they can take hours. The partial pruning in Fig.~\ref{fig:teaser}-B2 is performed online and each computation takes about 0.6 seconds on an Nvidia Titan RTX GPU. The head attention strengths and the tSNE layout for head attention patterns are both computed offline as they only need to be computed once and directly plugged into our system.
In terms of storage, the raw attention weights consume the most space, ranging from 11 GB to 178 GB, depending on the number of heads in the studied ViT. Other data (e.g., images, probabilities, tSNE results) take about 300 MB in total.

\textbf{\textit{Limitations and Future Work.}} Our head importance analysis relies on leave-one-out ablations, which do not consider the interaction between heads. In some cases, one head could be important only if another head is pruned. The analysis can be further extended to higher-order interactions, which is our planned future work. 
Second, our current analysis focuses on the attentions between two consecutive attention layers only. In the future, we would like to explore attention aggregation methods, e.g.,~\cite{abnar2020quantifying}, to interpret heads' impact across multiple layers.
Lastly, we plan to investigate if the head importance, head attention strengths, and head attention patterns show any class-specific or dataset-specific trends. This will help to diagnose class-related performance issues and validate our findings in more datasets.

\section{Conclusion}

In this paper, we introduce a visual analytics solution to interpret ViTs.
Our interpretation was carried out from three perspectives. First, we answer what heads are more important by introducing multiple head-importance metrics. Second, we explain why a head is important by disclosing its attention strength distribution across image patches in the 2D spatial context. Third, we adopt an unsupervised learning method to exhaustively summarize the possible attention patterns. Through concrete case studies conducted together with multiple experienced deep learning experts, we verify the efficacy of our visual interpretation solution.


%



\ifCLASSOPTIONcompsoc
  \section*{Acknowledgments}
\else
  \section*{Acknowledgment}
\fi
This research is sponsored in part by the National Institute of Health through grants 1R01CA270454-01 and 1R01CA273058-01.

\ifCLASSOPTIONcaptionsoff
  \newpage
\fi



\bibliographystyle{IEEEtran}
\bibliography{IEEEabrv,main}
%



%


\vspace{-0.2in}
\begin{IEEEbiographynophoto}{Yiran Li}
is a Ph.D. Candidate in computer science at the University of California, Davis. She received her B.S. degree in Mathematical Sciences from Zhejiang University in 2018. Her research interests are visual analytics and interpretable machine learning.
\end{IEEEbiographynophoto}

\vspace{-0.2in}
\begin{IEEEbiographynophoto}{Junpeng Wang}
is a research scientist at Visa Research. He received his B.E. degree in software engineering from Nankai University, M.S. degree in computer science from Virginia Tech, and Ph.D. degree in computer science from the Ohio State University. His research interests are broadly in visualization, visual analytics, and explainable AI.
\end{IEEEbiographynophoto}

\vspace{-0.2in}
\begin{IEEEbiographynophoto}{Xin Dai} is a staff research scientist at Visa Research. He received his B.E. and M.S. in computer science from Beijing Jiaotong University in 2012 and 2016, and Ph.D. in computer science from Worcester Polytechnic Institute in 2022. His research interests are in data mining and machine learning.
\end{IEEEbiographynophoto}

\vspace{-0.2in}
\begin{IEEEbiographynophoto}{Liang Wang}
is a principal research scientist at Visa Research.  His research interests are in data mining, machine learning, and fraud analytics. Liang received his Ph.D. degree in Computer Science from Facult\'e Polytechnique de Mons, Mons, Belgium with highest honors, and his BS degree in Electrical Engineering \& Automation and his MS degree in Systems Engineering, both from Tianjin University. Prior to joining Visa, He has worked in Yahoo!, eBay/PayPal, and FICO for bankcard fraud detection. He is the inventor of over 20 patents and has published over 30 papers in international journals and conferences.
\end{IEEEbiographynophoto}

\vspace{-0.2in}
\begin{IEEEbiographynophoto}{Chin-Chia Michael Yeh}
is a Staff Research Scientist at Visa Research. He received his Ph.D. in Computer Science from University of California, Riverside. His Ph.D. thesis "Toward a Near Universal Time Series Data Mining Tool: Introducing the Matrix Profile," received Doctoral Dissertation Award Honorable Mention at KDD 2019. He has published papers in top venues, including KDD, VLDB, ICDM and others. His research interests are in data mining, machine learning, and time series analysis.
\end{IEEEbiographynophoto}

\vspace{-0.2in}
\begin{IEEEbiographynophoto}{Yan Zheng} is currently a Senior Staff Research Scientist at Visa Research. Yan received her Ph.D. in Computer Science from University of Utah in 2017. She has published papers in top venues, including KDD, SIGMOD, ICDM and others. Her research interests are in data mining, machine learning, and representation learning.
\end{IEEEbiographynophoto}

\vspace{-0.2in}
\begin{IEEEbiographynophoto}{Wei Zhang}
is a principal research scientist and research manager at Visa Research and interested in big data modeling and advanced machine learning technologies for payment industry. Prior to joining Visa Research, Wei worked as a Research Scientist in Facebook, R\&D manager in Nuance Communications and also worked in IBM research over 10 years. Wei received his Bachelor and Master degrees from Department of Computer Science, Tsinghua University.
\end{IEEEbiographynophoto}

\vspace{-0.2in}
\begin{IEEEbiographynophoto}{Kwan-Liu Ma}
is a distinguished professor of computer science at the University of California, Davis. His research is in the intersection of data visualization, computer graphics, human-computer interaction, and high performance computing. For his significant research accomplishments, Ma received several recognitions including being elected as IEEE Fellow in 2012, recipient of the IEEE VGTC Visualization Technical Achievement Award in 2013, and inducted to IEEE Visualization Academy in 2019.
\end{IEEEbiographynophoto}







\end{document}